\documentclass{article}

\usepackage{arxiv}

\usepackage[utf8]{inputenc} % allow utf-8 input
\usepackage[T1]{fontenc}    % use 8-bit T1 fonts
\usepackage{hyperref}       % hyperlinks
\usepackage{url}            % simple URL typesetting
\usepackage{booktabs}       % professional-quality tables
\usepackage{amsfonts}       % blackboard math symbols
\usepackage{nicefrac}       % compact symbols for 1/2, etc.
\usepackage{microtype}      % microtypography
\usepackage{lipsum}		% Can be removed after putting your text content
\usepackage{graphicx}
\usepackage[numbers,sort&compress]{natbib}
\usepackage{doi}

\usepackage[printonlyused,nolist,withpage]{acronym}

\usepackage{csquotes}

\usepackage[version=4]{mhchem}

\usepackage{graphicx,float,booktabs,array,tabularx}
\usepackage{caption,subcaption, multirow, multicol}
\usepackage{multirow}      % Multi-row cells
\usepackage{threeparttable} % Table notes
\usepackage{xcolor}        % Colored text
\usepackage{amsmath}       % Math symbols
\usepackage{rotating}      % Rotated text
\usepackage{graphicx}      % Resizebox
\usepackage{pdflscape}       % For landscape option
\usepackage{siunitx}
\usepackage{makecell}

\usepackage{algorithm}
\usepackage{algpseudocode}

\usepackage[titletoc,title]{appendix}

\title{Parameter-Aware Ensemble SINDy for Interpretable Symbolic SGS Closure}

\author{ \href{https://orcid.org/0009-0000-4292-0655}{\includegraphics[scale=0.06]{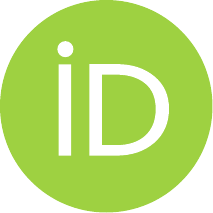}\hspace{1mm}Hanseul Kang}\thanks{Use footnote for providing further
		information about author (webpage, alternative
		address)---\emph{not} for acknowledging funding agencies.} \\
	Department of Mechanical Engineering\\
	Aalto University\\
	Espoo, Finland \\
	\texttt{hanseul.kang@aalto.fi} \\
	\And
	\href{https://orcid.org/0000-0001-6856-2200}{\includegraphics[scale=0.06]{orcid.pdf}\hspace{1mm}Ville Vuorinen} \\
	Department of Mechanical Engineering\\
	Aalto University\\
	Espoo, Finland \\
    \texttt{ville.vuorinen@aalto.fi} \\
	\And
	\href{http://orcid.org/0000-0003-0909-5969}{\includegraphics[scale=0.06]{orcid.pdf}\hspace{1mm}Shervin Karimkashi} \\
	Department of Mechanical Engineering\\
	Aalto University\\
	Espoo, Finland \\
	\texttt{shervin.karimkashiarani@aalto.fi} \\
}

\date{13 August 2025}

\renewcommand{\shorttitle}{}

\hypersetup{
  pdftitle={Parameter-Aware Ensemble SINDy for Interpretable Symbolic SGS Closure},
  pdfsubject={Symbolic regression, turbulence modelling, LES, SGS closure},
  pdfauthor={Hanseul Kang, Shervin Karimkashi, Ville Vuorinen},
  pdfkeywords={SINDy, subgrid-scale closure, turbulence modelling, symbolic regression, LES}
}

\begin{document}
\maketitle

% acronyms.tex — list all definitions in one place

\begin{acronym}[SINDy]

    \acro{DNS}{Direct Numerical Simulation}
    \acro{LES}{Large-Eddy Simulation}
    \acro{SGS}{Subgrid-Scale}
    \acro{SINDy}{Sparse Identification of Nonlinear Dynamic}
    \acro{MAE}{Mean Absolute Error}
    \acro{RMSE}{Root Mean Square Error}
    \acro{CPU}{Central Processing Unit}
    \acro{KdV}{Korteweg–de Vries}
    \acro{Burgers}{Burgers}
    \acro{CFD}{Computational Fluid Dynamic}
    \acro{DNN}{Deep Neural Network}
    \acro{PINN}{Physics-Informed Neural Network}
    \acro{ML}{Machine Learning}
    \acro{RANS}{Reynolds-averaged Navier–Stokes equation}
    \acro{PDE}{Partial Differential Equation}
    \acro{DSF}{Dimensional-Similarity Filter}
    \acro{ODE}{Ordinary differential equation}
    \acro{KdV}{Korteweg-de Vries}
    \acro{HPC}{High-performance computing}
    \acro{CFL}{Courant-Friedrichs-Lewy}
    \acro{LASSO}{Least Absolute Shrinkage and Selection Operator}
    \acro{CV}{Coefficient of Variation}
    \acro{WALE}{Wall-Adapting Local Eddy-viscosity}
    \acro{POD}{Proper Orthogonal Decomposition}
    \acro{GRF}{Gaussian Random Field}
    \acro{RL}{Reinforcement Learning}
    \acro{SpaRTA}{Sparse Regression of Turbulent Stress Anisotropy}
    \acro{AE}{Auto Encoder}
\end{acronym}

\begin{abstract}

This work designs a scalable, parameter-aware sparse regression framework for discovering interpretable partial differential equations and subgrid-scale closures from multi-parameter simulation data. Building on SINDy (Sparse Identification of Nonlinear Dynamics), the approach addresses key limitations through four enhancements. First, symbolic parameterisation enables physical parameters to vary within unified regression. Second, the Dimensional Similarity Filter enforces unit consistency while reducing candidate libraries. Third, memory-efficient Gram-matrix accumulation enables batch processing of large datasets. Fourth, ensemble consensus with coefficient stability analysis ensures robust model identification.

Validation on canonical one-dimensional benchmarks demonstrates consistent discovery of governing equations across parameter ranges. Applied to filtered Burgers datasets, the framework autonomously discovers the SGS closure $\tau_{\mathrm{SGS}} = 0.1604\cdot\Delta^2\left(\frac{\partial \bar{u}}{\partial x}\right)^2$ with the SINDy-discovered Smagorinsky constant $C_s^{\text{SINDy}} \approx 0.4005$ without predefined closure assumptions, recovering Smagorinsky-type structure directly from data.

The discovered model achieves $R^2 = 0.885$ across filter scales and demonstrates improved prediction accuracy compared to classical SGS closures. The ability of the framework to identify physically meaningful SGS forms and calibrate coefficients offers a complementary approach to existing turbulence modelling methods, contributing to the broader field of data-driven turbulence closure discovery.

\end{abstract}

% keywords can be removed
\section{Introduction}
% Motivation
% Context + Conventional ML with PINN
As computational performance grows and affordable simulations span ever-wider regimes, applications of \ac{ML} play a growing role in fluid mechanics~\cite{Raissi2019,Forootani2024,Zolman2024,Gao2025,Champion2019,Fukami2021,Beck2019,Meng2023}. While \acp{DNN} effectively approximate nonlinear solvers, their black-box nature limits interpretability and integration with established \ac{CFD} solvers. \acp{PINN} embed governing \acp{PDE} into the loss function, improving interpretability~\cite{Raissi2019}, yet they conceal the physical mechanisms truly governing the learned dynamics and require a priori knowledge of those equations.

%% SINDy and direction
By contrast, \ac{SINDy} constructs physics-informed libraries of candidate terms and uses sparse regression to reveal explicit governing laws, even under partial knowledge of the true form~\cite{Brunton2016,Rudy2017,Rudy2019,Messenger2021}. This work extends \ac{SINDy} with parameter-aware libraries and physics-constrained term selection to automatically discover interpretable \ac{SGS} closure models, eliminating manual parameter tuning while maintaining compatibility with existing CFD workflows.

% Literature Review
%% SINDy (Brunton)
Brunton et al.~\cite{Brunton2016} pioneered \ac{SINDy} and demonstrated that interpretable governing equations can be recovered from high-dimensional nonlinear systems for \ac{ODE} applications. The method constructs a feature library of candidate functions, such as polynomials and trigonometric modes, and applies sparsity-promoting regression to identify only the terms that truly govern the dynamics. This approach, however, faces a fundamental paradox. While it promises assumption-free discovery of the governing equation form, practitioners must still decide which terms from the candidate library to include or exclude. Conservative library construction based only on \textit{obvious} terms defeats the purpose of assumption-free discovery, while overly inclusive libraries lead to computational intractability and potential overfitting.

%% PDE-FIND
Rudy et al.~\cite{Rudy2017} built on this foundation and introduced the PDE-FIND algorithm to discover \acp{PDE} from field measurements by augmenting the library with spatial and temporal derivatives. The method flattens spatio-temporal data into large design matrices, each column representing a candidate function. PDE-FIND subsamples the data to reduce computational cost, which requires limiting a portion of high-fidelity experimental or computational datasets.

%% Parametric-SINDy
Current \ac{SINDy} assumes fixed coefficients, limiting its applicability to systems where governing laws vary across different physical parameter regimes. Rudy et al.~\cite{Rudy2019} addressed parameter-dependent discovery using Sequential Grouped Threshold Ridge Regression, demonstrating identification of \acp{PDE} with time-varying coefficients. While effective for discovering temporal parameter variation within individual trajectories, this group-sparsity approach requires preassembling the full feature library across all sampled conditions. This becomes computationally demanding when scaling to large parameter spaces.

%% ML-SGS Literature
Recent efforts to apply \ac{DNN} for \ac{SGS} closure modelling in \ac{LES} have made notable progress whilst revealing key limitations that motivate alternative approaches~\cite{Meng2023, Kurz2023, Beck2019}. Kurz et al.~\cite{Kurz2023} successfully employed \ac{RL} to learn spatially and temporally adaptive Smagorinsky coefficients for implicitly filtered LES of homogeneous isotropic turbulence, demonstrating stable long-term simulations that outperform established analytical models. However, the approach assumes the continued validity of the underlying Smagorinsky framework and requires substantial computational resources due to the iterative nature of \ac{RL} training with multiple \ac{LES} simulations~\cite{Sutton2018}. Meng et al.~\cite{Meng2023} effectively demonstrated \acp{DNN} for \ac{SGS} closure modelling in compressible turbulent channel flow across varying Mach and Reynolds numbers, achieving correlation coefficients larger than 0.91 and outperforming traditional models. Their approach, though successful, requires dual network architectures with separate networks for SGS stress and heat flux and relies on single-point velocity and temperature gradients as inputs, potentially limiting transferability to more complex flow configurations compared to approaches that incorporate broader spatial information.

% Research Gap
Despite significant advances, existing approaches exhibit practical challenges when applied to complex fluid problems. They face library-construction ambiguity where practitioners must subjectively decide which terms to include. This underscores the advantage of \ac{SINDy} with its assumption-free feature. Computational scalability forces algorithms like PDE-FIND to sacrifice portions of valuable data through subsampling~\cite{Rudy2017}. 
Most studies also remain limited to fixed parameter values~\cite{Brunton2016,Rudy2017,Gao2025}. Yet fluid phenomena behave fundamentally differently across parameter regimes.

Classical \ac{SGS} closures like the Smagorinsky model demand manual tuning of model constants with limited guidance on optimal values for specific conditions~\cite{SMAGORINSKY1963}. While \ac{DNN}-based approaches demonstrate improved accuracy, they exhibit limited interpretability, complex integration requirements with established \ac{CFD} solvers, and computational overhead that can destabilise iterative convergence. 

The potential of \ac{SINDy} for discovering symbolic closure models remains underexplored and offers an opportunity to bridge the gap between empirical models and black-box \acp{DNN}. The data-driven symbolic approach created here provides both accuracy and physical insight while automatically discovering interpretable expressions with plug-and-play compatibility for existing \ac{CFD} workflows.

%  Objectives
To address these limitations, this work pursues four aims:
\begin{enumerate}
  \item Parameter-aware \ac{SINDy} libraries: Develop feature libraries whose terms explicitly depend on physical parameters (viscosity $\nu$, filter width $\Delta$), enabling single models to generalise across operating conditions.
  \item Memory-efficient sparse regression: Implement Gram-matrix accumulation to avoid full feature-matrix assembly, eliminating memory bottlenecks while preserving complete datasets.
  \item Physics-constrained term selection: Apply \ac{DSF} to eliminate nonphysical candidates before regression, reducing computational waste and improving physical consistency.
  \item Symbolic \ac{SGS} closure model discovery: Extend \ac{SINDy} to identify interpretable \ac{SGS} closure models for \ac{LES}, providing closed-form alternatives to black-box \ac{DNN} approaches.
\end{enumerate}

To achieve these objectives, we validate the framework across three canonical one-dimensional test cases involving the Heat, Burgers', and \ac{KdV}-Burgers' equations. This deliberate focus on 1D preserves the interpretability that motivates this work. Higher-dimensional extensions typically require coupling \ac{SINDy} with black-box dimensionality reduction models such as autoencoders~\cite{Champion2019, Conti2023}. These models compromise the symbolic transparency we seek to maintain. The 1D setting provides an ideal environment for validating our interpretable discovery pipeline. Heat diffusion and Burgers' equation cases confirm the effectiveness of memory-efficient, parameter-aware libraries. They also validate automated term selection with \ac{DSF}. The \ac{KdV}-Burgers' equation demonstrates the framework's capability to capture complex nonlinear dynamics. Meanwhile, \ac{SGS} stress regression reveals interpretable closure discovery from Burgers-based data.
\section{Methodology}

\subsection{Case Studies}

This work considers three canonical \acp{PDE}, where $u(x,t)$ represents the field variable, $u_t = \partial u/\partial t$ denotes the temporal derivative, and $u_x = \partial u/\partial x$, $u_{xx} = \partial^2 u/\partial x^2$, and $u_{xxx} = \partial^3 u/\partial x^3$ represent the first, second, and third spatial derivatives, respectively.

The heat-diffusion equation represents pure diffusion:
\begin{align}
    u_t = \nu u_{xx},
    \label{eq:heat_diffusion}
\end{align}
where $\nu$ denotes the kinematic viscosity.

Burgers' equation combines nonlinear advection with diffusion~\cite{BATEMAN1915,Burgers1948}:
\begin{align}
    u_t + uu_x = \nu u_{xx}.
    \label{eq:burgers}
\end{align}

The \ac{KdV} equation models waves on shallow water surfaces~\cite{Gao2025}:
\begin{align}
    u_t + uu_x + u_{xxx} = 0.
    \label{eq:kdv}
\end{align}
To incorporate viscous effects and introduce multiple governing parameters for systematic algorithm evaluation, this work extends Equation~\ref{eq:kdv}:
\begin{align}
    u_t + C_1 uu_x + C_2 u_{xxx} = \nu u_{xx},
    \label{eq:kdv_burgers}
\end{align}
where $C_1$ and $C_2$ serve as tunable parameters that enable systematic evaluation of discovery algorithms under varying equation coefficients.

\begin{figure}[htbp]
    \centering
    \begin{subfigure}[b]{0.325\textwidth}
        \centering
        \includegraphics[width=\textwidth]{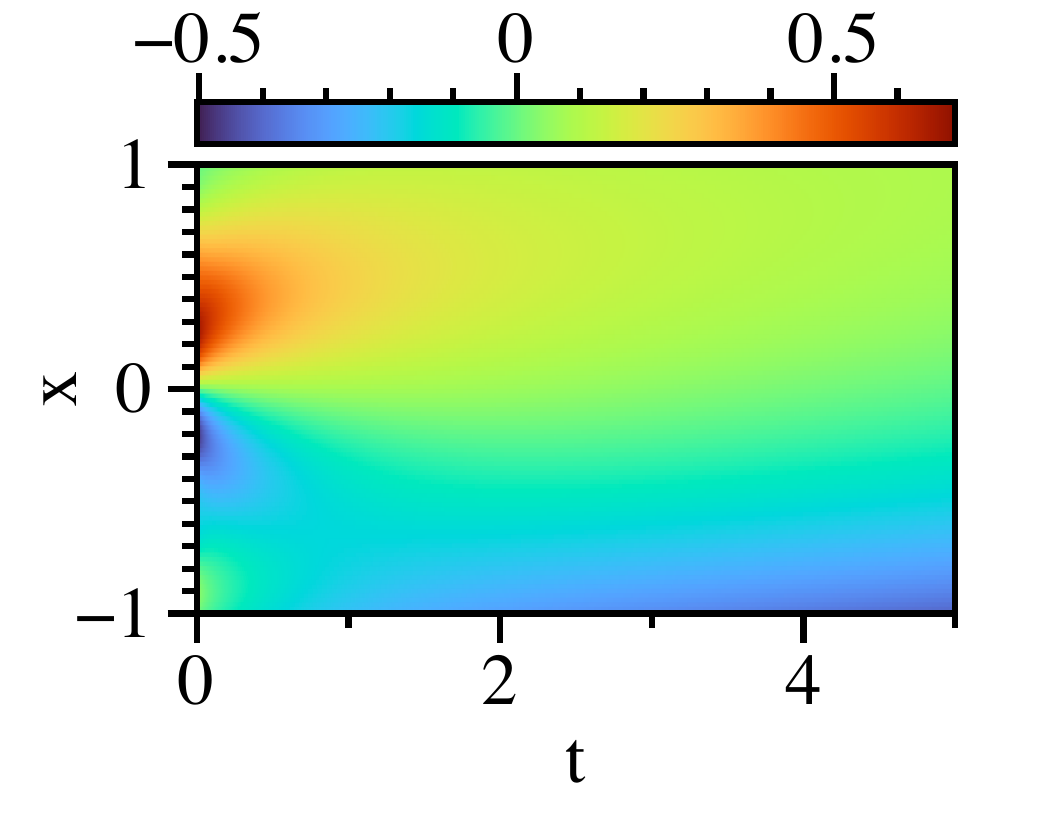}
        \caption{Heat diffusion}
        \label{fig:heat_diffusion}
    \end{subfigure}
    \hfill
    \begin{subfigure}[b]{0.325\textwidth}
        \centering
        \includegraphics[width=\textwidth]{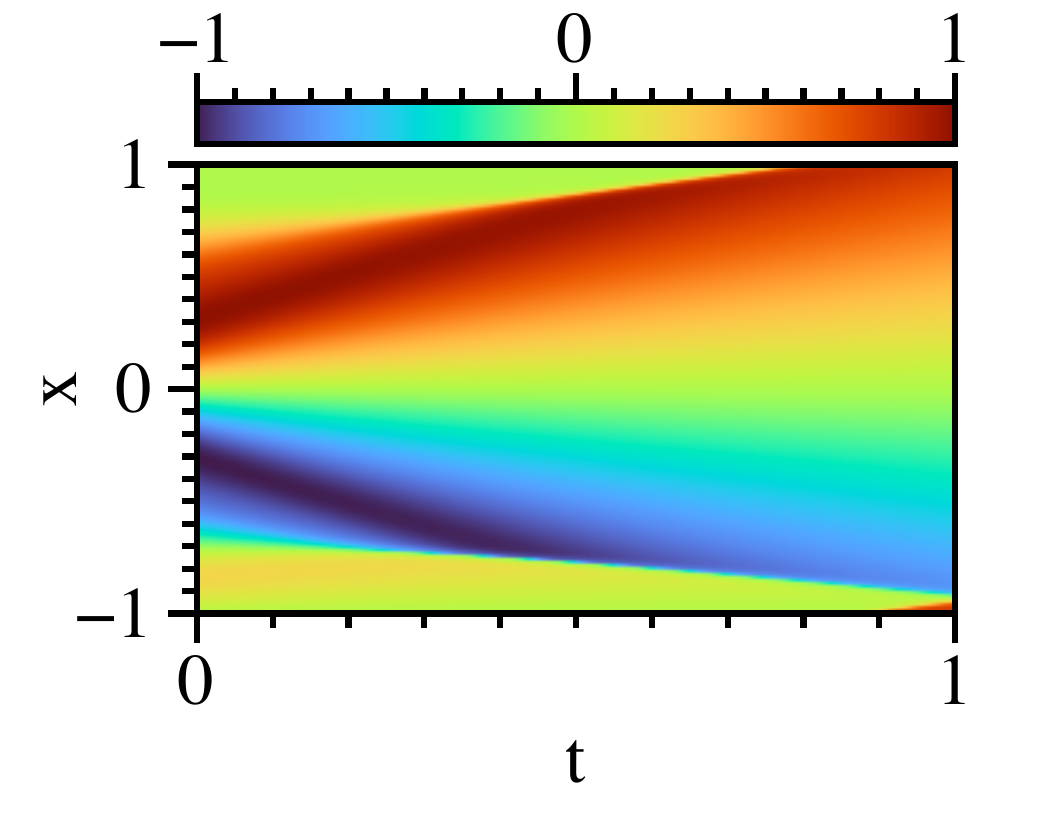}
        \caption{Burgers' equation}
        \label{fig:burgers_equation}
    \end{subfigure}
    \hfill
    \begin{subfigure}[b]{0.325\textwidth}
        \centering
        \includegraphics[width=\textwidth]{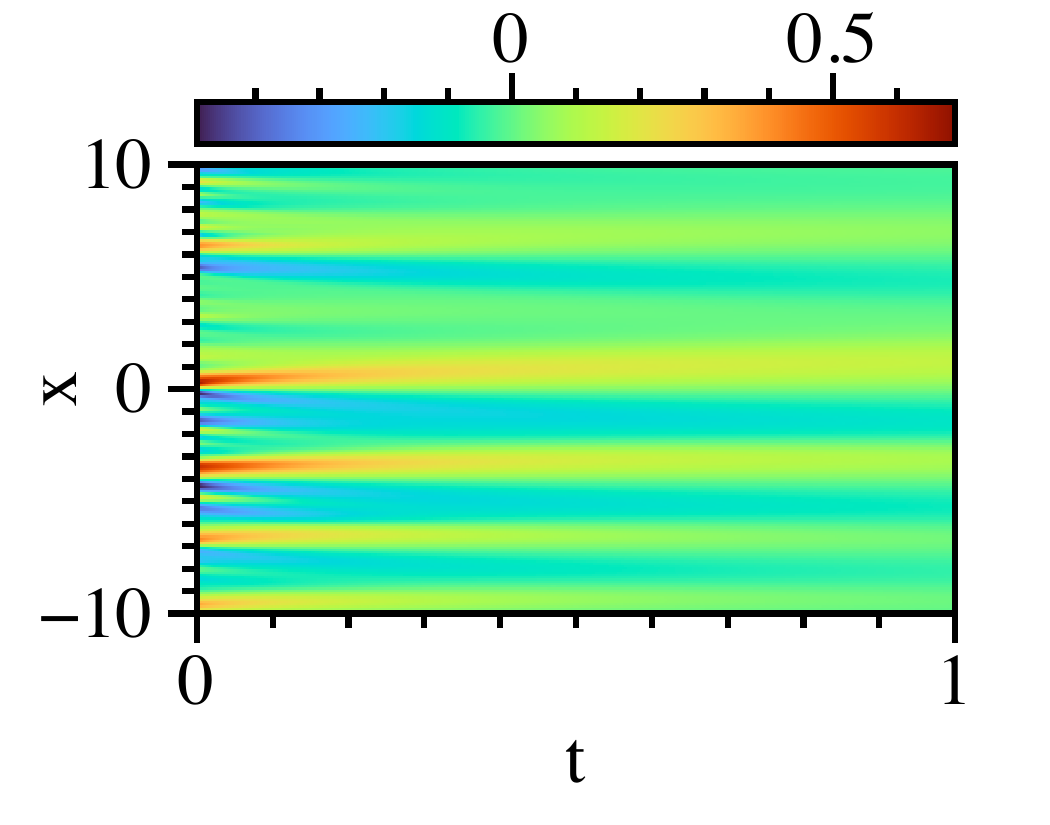}
        \caption{KdV-Burgers equation}
        \label{fig:kdv_burgers}
    \end{subfigure}
    \caption{Representative solution behaviour of the three \acp{PDE} with Perlin noise initial conditions. Viscosity values: $\nu = 8.36 \times 10^{-2}$ (heat diffusion), $\nu = 2.66 \times 10^{-3}$ (Burgers), $\nu = 3.76 \times 10^{-1}$ (KdV-Burgers).}
    \label{fig:pde_comparison}
\end{figure}

Finite difference methods generate datasets by numerically integrating the \acp{PDE}. One-dimensional Perlin noise provides the initial conditions, sampled as smooth, spatially correlated random fields~\cite{Perlin1985}. The initial condition facilitates capturing relevant physical phenomena across different equation types. Figure~\ref{fig:pde_comparison} shows one training case.

\begin{table}[htbp]
  \centering
  \caption{Simulation setup for the three 1D case studies.}
  \label{tab:case_studies_setup}
  \begin{threeparttable}
  \small
  \begin{tabular}{@{}lcccc@{}}
    \toprule
    \textbf{Equation} & \textbf{Domain $(x,t)$} & \textbf{Viscosity $\nu$} & \textbf{Grid $(N_x, N_t)$} & \textbf{Initial Condition\tnote{a}} \\
    \midrule
    Heat-diffusion    & $[-1,1]\times[0,5]$           & $[0.01,0.1]$            & $(100,3000)$            & Perlin noise      \\
    Burgers'          & $[-1,1]\times[0,1]$           & $[0.001/\pi,0.01/\pi]$  & $(150,500)$             & Perlin noise      \\
    KdV--Burgers'      & $[-10,10]\times[0,3]$         & $[0.045,0.45]$         & $(150,500)$             & Perlin noise      \\
    \bottomrule
  \end{tabular}
  \begin{tablenotes}[flushleft]
  \small
  \item[a] Perlin noise parameters: octaves $\in [1,3]$, frequency $\in [0.5,1.5]$.
  \end{tablenotes}
  \end{threeparttable}
\end{table}

\subsection{Explicit Filtering and SGS Closures}
\label{subsec:sgs_stress}

This study applies \ac{LES} methodology to Burgers' equation, which contains both nonlinear advection and viscous diffusion and exhibits shock formation. In addition, prior studies have applied \ac{SGS} closure modelling to Burgers' equation~\cite{Li2016,Peng2009}, motivating its adoption as a benchmark for testing the proposed symbolic discovery pipeline.

A naive application of spatial filtering to Burgers' equation yields
\begin{align}
  \bar u_t+\bar u\,\bar u_x=\nu \bar u_{xx},
\label{eq:naive}
\end{align}
However, due to the nonlinearity of the convective term, $\overline{u\,u}\neq\bar u\,\bar u$, necessitating the introduction of a \ac{SGS} stress term 
$\tau_{SGS}\equiv\overline{u^2}-\bar u^{\,2}$~\cite{Pope2000}.  
The correct filtered equation becomes
\begin{align}
  \bar u_t+\bar u\,\bar u_x+\frac{\partial \tau_{SGS}}{\partial x}=\nu \bar u_{xx}.  \label{eq:filtered}
\end{align}

The filtered velocity emerges as the convolution
\begin{align}
  \bar u(x)=\int_{-\infty}^{\infty}G_\Delta(x-r)\,u(r)\,dr,
\end{align}
where the kernel for a box filter of width \(\Delta\) equals  
$G_\Delta(\eta)=\tfrac1\Delta$ if $|\eta|\le\Delta/2$, and equals $0$ otherwise.  Hence
\begin{align}
  \bar u(x)=\frac1\Delta\int_{x-\Delta/2}^{x+\Delta/2}u(r)\,dr.
\end{align}

The corresponding SGS stress becomes
\begin{align}
  \tau_{SGS}(x)=\overline{u^{2}}-\bar u^{\,2}
  =\frac1\Delta\int_{x-\Delta/2}^{x+\Delta/2}u^{2}(r)\,dr
   -\left(\frac1\Delta\int_{x-\Delta/2}^{x+\Delta/2}u(r)\,dr\right)^{2}.
\end{align}

Several analytical approaches exist for approximating the SGS stress. The Taylor expansion around $x$ yields the leading-order theoretical closure
\begin{align}
  \tau_{SGS}\;\approx\;\frac{\Delta^{2}}{12}\bigl(u_x^{2}-u\,u_{xx}\bigr).
\label{eq:taylor_sgs}
\end{align}

Alternatively, the Leonard expansion applied to the \ac{SGS} stress in 1-D~\cite{Leonard1975, Peng2009} provides a systematic series:
\begin{align}
    \tau_{SGS} \approx \Delta^2\bar{u}_{x}^2 + \frac{\Delta^4}{2!}\bar{u}_{xx}^2 + \frac{\Delta^6}{3!}\bar{u}_{xxx}^2 + \cdots.  
\label{eq:leonard_sgs}
\end{align}

One widely adopted SGS closure model follows the Smagorinsky model~\cite{SMAGORINSKY1963}, which approximates the SGS stress as proportional to the square of the local strain rate magnitude. For the 1-D Burgers' equation, this model takes the form~\cite{Li2016}:
\begin{align}
    \tau_{SGS} =  -C_s^2 \Delta^2 \bar{u}_x |\bar{u}_x|,
\label{eq:smagorinsky_sgs}
\end{align}
where $C_s$ represents the Smagorinsky constant. The value of $C_s$ has been empirically determined to range between 0.1 and 0.24~\cite{Wilcox2006}. However, the optimal value remains problem-dependent and requires manual calibration for different flow configurations.

Throughout this work, the analysis refers to the Taylor-expanded and Leonard-expanded \ac{SGS} closures as \textit{Taylor SGS} and \textit{Leonard SGS}, respectively. This maintains clarity and avoids conceptual ambiguity. These analytical closures serve as primary benchmarks because they provide explicit, symbolic forms. This permits direct comparison of the discovered physical structure against established theory. The approach aligns with the principal goal of the \ac{SINDy} framework, which targets interpretable symbolic discovery. A direct performance comparison with opaque models such as \acp{DNN} would fail to serve this purpose.

\subsection{SINDy Library Construction}

The \ac{SINDy} library $\Theta$ forms the foundation for discovering PDE terms. Construction follows a systematic multi-step process that ensures physical consistency and mathematical rigour.

Base terms constitute irreducible building blocks:
\begin{align}
   & \Theta_{\text{base}} = \Theta_{\text{const}} \cup \Theta_{\text{field}} \cup \Theta_{\text{param}} \cup \Theta_{\text{deriv}}, \\
    &\Theta_{\text{const}} = \{1\}, \\
    &\Theta_{\text{field}} = \{u\}, \\
    &\Theta_{\text{param}} = \{p_1, p_1^{-1}, p_2, p_2^{-1}, \ldots, p_P, p_P^{-1}\}, \\
    &\Theta_{\text{deriv}} = \{u_x|_{\text{cd1}}, u_x|_{\text{uw2}}, u_{xx}|_{\text{cd2}}, \ldots\},
\end{align}
where subscripts denote term categories of constants, field variables, physical parameters, and spatial derivatives. The parameter set contains $P$ physical quantities with their inverses for flexible coefficient scaling, such as kinematic viscosity $p_1 = \nu$, filter width $p_2 = \Delta$, and equation coefficients $p_3 = C_1$. Derivative notation $u_x|_{\text{scheme}}$ distinguishes finite difference methods, as first-order central difference \texttt{cd1}, second-order central difference \texttt{cd2}, and second-order upwind \texttt{uw2}. Higher-order derivatives $u_{xxx}$, $u_{xxxx}$ can be added for dispersive or complex phenomena as needed.

For degree $d$, the framework generates all possible monomial combinations:
\begin{align}
    \Theta_{\text{raw}}^{(d)} = \left\{ \prod_{i=1}^{|\Theta_{\text{base}}|} \theta_i^{k_i} : \sum_{i=1}^{|\Theta_{\text{base}}|} k_i = d, \, k_i \geq 0 \right\},
\end{align}
where $\theta_i \in \Theta_{\text{base}}$ and $k_i$ represent non-negative integer exponents.

The following reduction rules apply sequentially:

\begin{enumerate}
    \item Inverse Cancellation: For each parameter $p_j$:
    \begin{align}
        p_j^a \cdot (p_j^{-1})^b \rightarrow \begin{cases}
            p_j^{a-b} & \text{if } a > b \\
            (p_j^{-1})^{b-a} & \text{if } b > a \\
            1 & \text{if } a = b.
        \end{cases}
    \end{align}
    
    \item Commutative Ordering: Order terms canonically to eliminate duplicates.
    
    \item Pure Parameter Elimination: Remove monomials containing only parameters and their inverses, excluding field variables or derivatives.
\end{enumerate}

The raw number of degree-$d$ monomials equals $\binom{B + d - 1}{d}$, where $B = |\Theta_{\text{base}}|$. Rules 1--3 remove inverse cancellations and pure-parameter products. The surviving set requires algorithmic computation because a closed-form count becomes cumbersome when multiple parameters and their inverses interact within the same monomial.

The total library size up to the maximum degree $D$ becomes:
\begin{align}
    |\Theta_{\text{total}}| = \sum_{d=0}^{D} |\Theta_{\text{pruned}}^{(d)}|,
\end{align}
where $|\Theta_{\text{pruned}}^{(d)}|$ denotes the number of terms surviving the reduction process at degree $d$.

This combinatorial framework scales to arbitrary parameter numbers while preserving physically meaningful terms and eliminating mathematically redundant combinations. However, the minimal intervention approach generates numerous terms with inconsistent dimensions and unclear physical interpretation that can lead to overfitting in sparse regression. The library size also grows rapidly with degree and parameter count, necessitating further filtering strategies.

\subsection{Dimensional Similarity Filter (DSF)}

\begin{figure}[!htbp]
    \centering
    \includegraphics[width=0.45\textwidth]{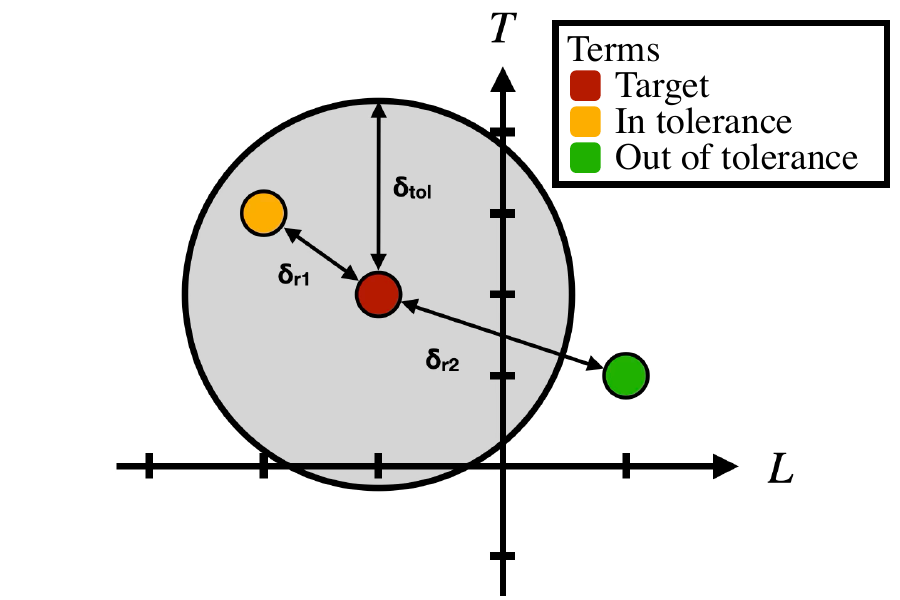}
    \caption{\ac{DSF} concept in 2D $L$-$T$ space. 
            The red circle represents the target term $u_t$, 
            the yellow circle denotes a candidate term within tolerance ($\delta_{r1} \leq \delta_{\text{tol}}$), 
            and the green circle marks a rejected term outside tolerance ($\delta_{r2} > \delta_{\text{tol}}$). 
            The vertical line at $\delta_{\text{tol}}$ indicates the filtering boundary.}
    \label{fig:dsf_filter}
\end{figure}

The raw degree-based library $\Theta_{\text{deg}}$ can grow combinatorially, making direct regression computationally prohibitive. To address this challenge, the framework develops the \ac{DSF} that applies a quick dimensional consistency check before any regression.

Following the principle of dimensional homogeneity~\cite{Fourier1955}, every primitive quantity receives a dimension vector $\mathbf d=[L,T]$, as in Table~\ref{tab:dimensional_analysis}. The vector for a composite term equals the sum of its constituents.

Recent deep symbolic regression approaches enforce unit consistency during expression generation by applying local unit constraints at each step, ensuring that only dimensionally balanced expressions can be formed~\cite{Tenachi2023}. By contrast, \ac{DSF} acts earlier in the pipeline, filtering a pre-assembled, parameter-aware SINDy library. It removes terms incompatible with the target operator employing a hard or soft tolerance on distance in the dimension space. This pre-regression screening curbs combinatorial growth and improves numerical stability while leaving the subsequent sparse regression unchanged.

If strict dimensional consistency proves required, the analysis applies a \textit{hard filter} where only exact matches to the target vector $\mathbf d^{\star}$ remain,
\begin{align}
    \Theta_{\text{hard}}=\{\theta\in\Theta_{\text{deg}}\mid\mathbf d_\theta=\mathbf d^{\star}\}.
\end{align}

For broader searches, the framework implements a \textit{soft filter} that allows a tolerance band around exact dimensional matches. Let $\mathbf{d}^{\star} = [d^{\star}_L, d^{\star}_T]$ denote the length and time dimensions of the target derivative (e.g., $\partial u/\partial t$), and $\mathbf{d}_\theta = [d^{\,L}_\theta, d^{\,T}_\theta]$ denote the corresponding dimensions of a candidate library term $\theta$. 
Normalised component-wise distances
\begin{align}
    d_L=\frac{|d^{\star}_L-d^{\,L}_\theta|}{\max_L},\quad
    d_T=\frac{|d^{\star}_T-d^{\,T}_\theta|}{\max_T},
\end{align}
combine through either the taxicab or Euclidean metric,
\begin{align}
\delta_{\mathrm{taxicab}}=\frac{1}{2}(d_L+d_T),\qquad
\delta_{\mathrm{euclid}}=\sqrt{\frac{d_L^{2}+d_T^{2}}{2}}.
\end{align}
The normalisation factors $\max_L$ and $\max_T$ represent the maximum absolute dimensional values across all candidate terms, ensuring scale-invariant distance computation. Terms with $\delta < \delta_{\mathrm{tol}}$ remain in the filtered library, where $\delta_{\mathrm{tol}}$ represents a user-defined tolerance.

Terms satisfying $\delta \le \delta_{\mathrm{tol}}$ form the reduced set
\begin{align}
    \Theta_{\text{DSF}}=\bigl\{\theta\in\Theta_{\text{deg}}\mid
\delta(\mathbf{d}_\theta,\mathbf{d}^{\star}) \le \delta_{\mathrm{tol}}\bigr\}.
\end{align}

Because the test remains purely algebraic, computational cost scales linearly with
$|\Theta_{\text{deg}}|$ while typically reducing the library size by
one to two orders of magnitude, as shown in Figure~\ref{fig:dsf_filter}.

\begin{table}[htbp]
    \centering
    \caption{Dimensional analysis of common terms in the \ac{SINDy} library.}
    \label{tab:dimensional_analysis}
    \small
    \begin{tabular}{@{}lcc@{}}
        \toprule
        \textbf{Term} & \textbf{Dimension $[L, T]$} & \textbf{Physical Meaning} \\
        \midrule
        $1$ & $[0, 0]$ & Dimensionless constant \\
        $u$ & $[1, -1]$ & Velocity \\
        $u_x|_{cd1}, u_x|_{uw2}$ & $[0, -1]$ & Velocity gradient \\
        $u_{xx}|_{cd2}$ & $[-1, -1]$ & Velocity curvature \\
        $u_{xxx}|_{cd2}$ & $[-2, -1]$ & Third-order spatial derivative \\
        $\nu$ & $[2, -1]$ & Kinematic viscosity \\
        $C_1$ & $[0, 0]$ & Nonlinear advection coefficient \\
        $C_2$ & $[3, -1]$ & Dispersive scaling coefficient \\
        \bottomrule
    \end{tabular}
\end{table}

\subsection{Memory-Efficient Sparse Regression}

Large-scale \ac{SINDy} problems encounter significant storage demands when the library matrix $\Theta$ grows beyond practical limits. Current methods apply \ac{POD} or autoencoder to mitigate explosive growth~\cite{Champion2019, Russo2025, Conti2023}, enabling analysis of 2D and 3D cases but sacrificing physical interpretability.

The incremental accumulation of Gram matrices represents a well-known technique in numerical linear algebra~\cite{Horn2013}. Although Messenger et al.~\cite{Messenger2021} introduced Gram matrices to SINDy through their Weak SINDy formulation, our work employs them in the strong-form setting. The strong-form requires only the Gram pair $\mathbf{G} = \Theta^T\Theta$ and $\mathbf{b} = \Theta^T\mathbf{u}_t$. By partitioning $\Theta$ into batches and computing:
\begin{align}
    \mathbf{G} &= \sum_{i=1}^{B} \Theta_i^T \Theta_i, \\
    \mathbf{b} &= \sum_{i=1}^{B} \Theta_i^T \mathbf{u}_{t,i},
\end{align}
the footprint reduces from $\mathcal{O}(np)$ to $\mathcal{O}(p^2)$ without requiring \ac{POD}. This enables batch processing of arbitrarily large datasets using conventional \ac{SINDy} library construction while achieving equivalent reductions. The key insight emerges that most sparse regression algorithms only need $\mathbf{G}$ and $\mathbf{b}$, bypassing the original matrix $\Theta$~\cite{Brunton2016, Rudy2017}.

Once accumulated, any Gram-compatible solver transforms the regression problem to normal equations $\mathbf{G}\xi = \mathbf{b}$. Sparse regression methods (LASSO, ElasticNet, SR3) operate directly on this reduced representation~\cite{Hastie2009,Tibshirani2010}, enforcing sparsity via the $\ell_1$ norm as introduced by Rudy et al.~\cite{Rudy2017}.

\subsection{Ensemble Methods for Robust Discovery}

\ac{SINDy} relies on finite difference approximations, making it sensitive to measurement noise and boundary effects. To enhance robustness while discovering equations across cases with varying properties, we employ a bagging ensemble that leverages the symbolic nature of \ac{SINDy}.

Unlike typical \ac{ML} ensembles, the symbolic structure enables coefficient stability and term consensus analysis. We define two aggregation metrics:

First, \ac{CV} measures coefficient stability:
\begin{align}
    \mathrm{CV}_j = \frac{\sigma_j}{|\mu_j| + \epsilon},
\end{align}
where $\mu_j$ and $\sigma_j$ denote the mean and standard deviation of coefficient $j$ over the selected subset of estimators, and $\epsilon = 10^{-10}$ prevents division by zero.

Second, Consensus Frequency measures term appearance across bootstrap samples:
\begin{align}
    f_j = \frac{1}{M}\sum_{m=1}^{M} \mathbf{1}[|\xi_{j,m}| > \lambda_{\text{noise}}],
\end{align}
where $M$ denotes the number of estimators and $\mathbf{1}[\cdot]$ represents the indicator function. Setting $\lambda_{\text{noise}} = 10^{-5}$ excludes likely numerical noise from consensus counts. Physical parameters in the library produce coefficients of meaningful scale, so smaller values indicate artefacts rather than dynamics.

We align symbolic terms across estimators by name and define the selected set $S_j=\{\,m\in\{1,\ldots,M\}\mid |\xi_{j,m}|>\lambda_{\text{noise}}\,\}$. The consensus frequency equals $f_j=|S_j|/M$, with $\mu_j$ and $\sigma_j$ computed over $S_j$ only.

The method retains terms satisfying both $f_j \geq f_{\text{threshold}}$ and $\mathrm{CV}_j \leq \mathrm{CV}_{\text{threshold}}$, ensuring consistent selection and stable estimation. This dual-threshold approach enhances robustness against noise while preserving the interpretable symbolic equations of \ac{SINDy}.

% Example
%\xi_{j,1} = 0.12
%\xi_{j,2} = 0.00
%\xi_{j,3} = 0.15
%\xi_{j,4} = -0.20
%\xi_{j,5} = 0.00001
% S_j = \{1,3,4\}. Due to \lambda_{\text{noise}=1e-5} 
% f_j = \frac{|S_j|}{M} = \frac{3}{5} = 0.6
% \mu_j = \frac{0.12 + 0.15 + (-0.20)}{3} = 0.0233
 \section{Results and Discussion}
 \label{sec:result}
Treating boundaries in finite-difference schemes requires special consideration as it impacts numerical accuracy~\cite{Ferziger2020}. Consequently, all analyses in Section \ref{sec:result} employ a 2-cell wall buffer, excluding the outermost data points to mitigate this potential inaccuracy.
 
\subsection{Parameter-aware PDE identification}
\label{subsec:parameter_sensitive}
% METHOD: DSF + Gram accumulation + Ensemble consensus
% TASK: Recover classical PDEs (Heat, Burgers', KdV-Burgers) across parameter ranges
% SHOW: Exact coefficient recovery (ν∈[0.001,0.1], C₁∈[1,15], C₂∈[0.002,0.042])
% METRICS: |coeff_true - coeff_discovered|, consensus frequency, coefficient CV
% DEMONSTRATE: Complete pipeline handles parameter uncertainty in known physics

Section \ref{subsec:parameter_sensitive} validates the complete pipeline, including \ac{DSF}, Gram accumulation, and ensemble methods, on three \acp{PDE} with known analytical forms. The validation demonstrates exact coefficient identification across varying parameter ranges, addressing a key challenge for traditional \ac{SINDy} methods that assume fixed coefficients.

Heat and Burgers’ equations consider 17 candidate terms, while KdV–Burgers starts with 124 terms after applying DSF with tolerance 0.5. The framework employs polynomial libraries of degree 2 and 3, respectively. The ensemble method employs 10 estimators with 80\% data subsampling, consensus threshold 0.8, and coefficient variation threshold 0.15.

\begin{figure}[htbp]
    \centering
    \begin{subfigure}[t]{0.49\textwidth}
    \centering
        \includegraphics[width=\textwidth]{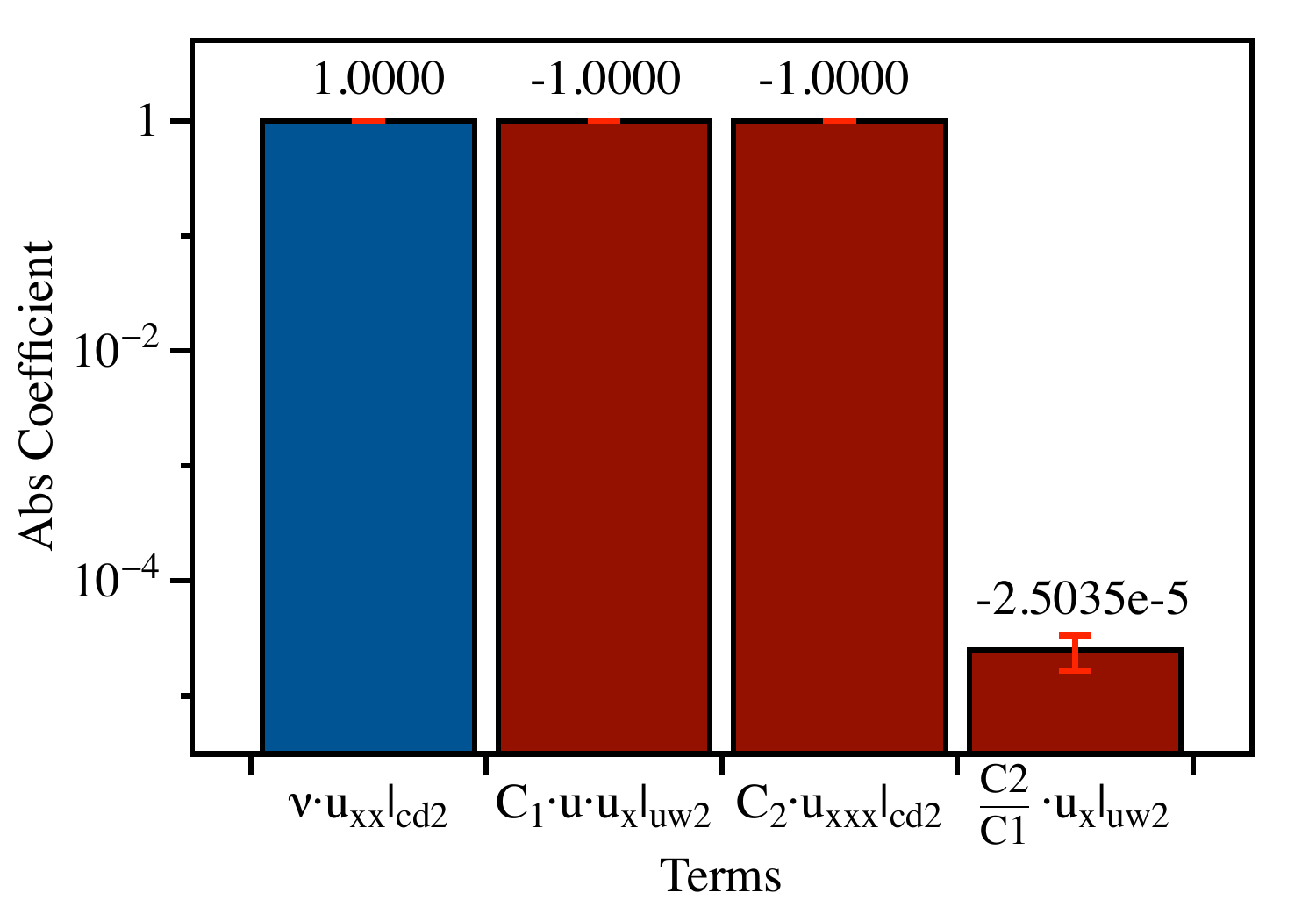}
        \caption{Candidate terms after \ac{SINDy}. Only stable terms $C_1 u u_x|_{uw2}$, $C_2 u_{xxx}|_{cd2}$, and $\nu u_{xx}|_{cd2}$ qualified our thresholds, with one additional term shown for illustration.}
        \label{fig:parameter_sensitive_trial1}
        \end{subfigure}
        \hfill
        \begin{subfigure}[t]{0.49\textwidth}
    \centering
        \includegraphics[width=\textwidth]{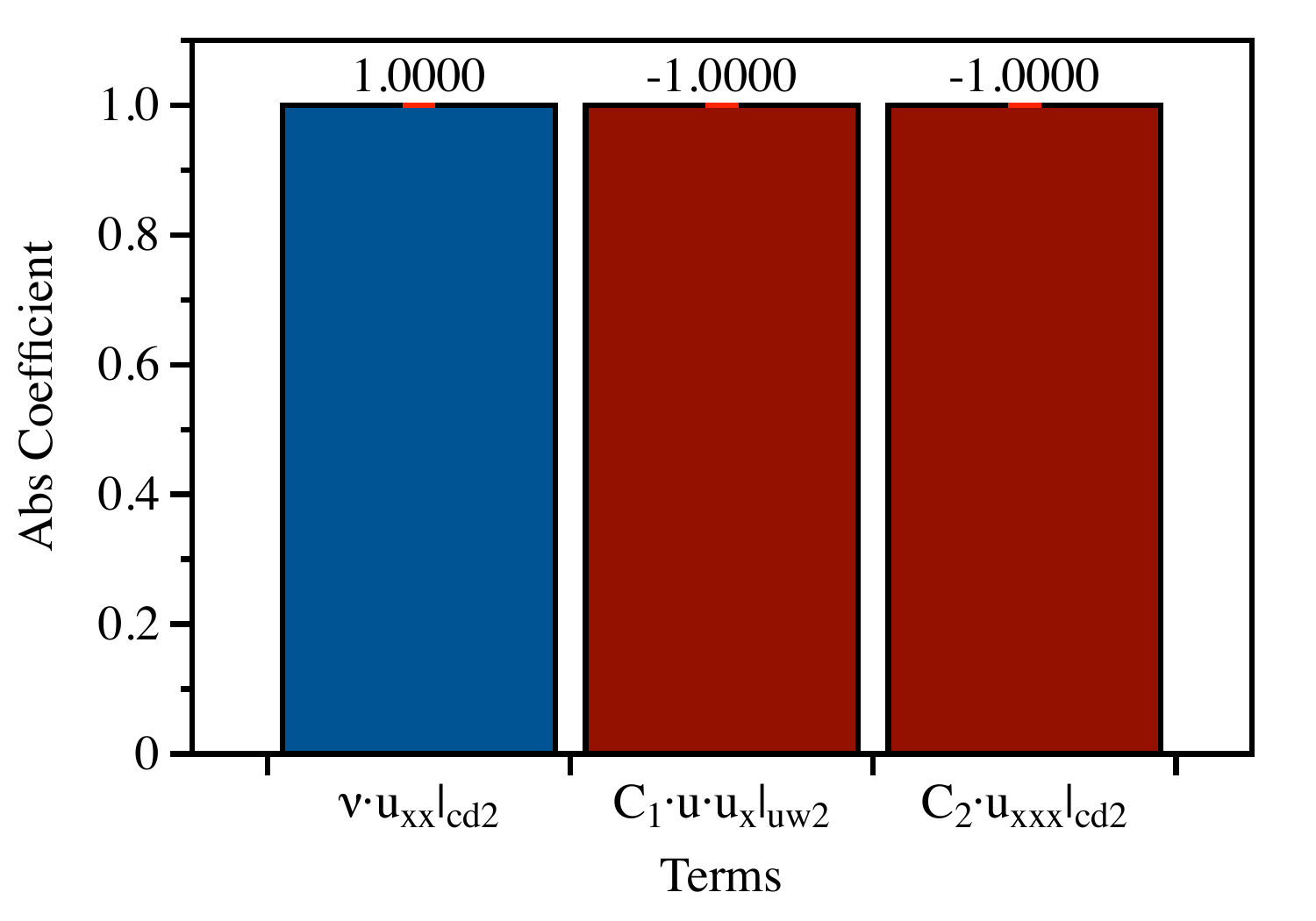}
        \caption{Final terms after refined regression using only qualified stable terms}
        \label{fig:parameter_sensitive_trial2}
        \end{subfigure}
    \caption{Two-stage filtering for \ac{KdV}-Burgers identification. Initial ensemble identifies candidate terms with only three terms meeting stability thresholds, shown in subfigure a. Refined regression using qualified terms produces exact coefficients with reduced variance, shown in subfigure b. Blue bars indicate positive coefficients and red bars indicate negative coefficients.}
\label{fig:parameter_sensitive}
\end{figure}

Figure~\ref{fig:parameter_sensitive_trial1} demonstrates successful PDE identification for the \ac{KdV}-Burgers equation employing the simulation setup detailed in Table~\ref{tab:case_studies_setup}. The SINDy regression identifies the three physically correct terms: the nonlinear convection term $C_1 u u_x|_{uw2}$, the third-order dispersion term $C_2 u_{xxx}|_{cd2}$, and the second-order diffusion term $\nu u_{xx}|_{cd2}$, which precisely match the target equation Equation~\ref{eq:kdv_burgers}. However, this regression employed the full candidate library containing additional terms, introducing coefficient variance during the sparse regression process.

To address this limitation, the analysis performs refined regression employing only the stable terms that qualified both CV and consensus thresholds. Figure~\ref{fig:parameter_sensitive_trial2} presents results from this two-stage approach, where only the physically meaningful terms that passed stability criteria remain for regression. This refined approach improves coefficient stability by regressing only the threshold-satisfied terms through iterative refinement. For the KdV–Burgers' case, refining the regression to the three retained terms reduces coefficient variations from the order of $10^{-7}$ in the initial full-library regression to the order of $10^{-8}$ in the final targeted regression.

Table~\ref{tab:pde_recovery} presents the ensemble PDE identification results representing the outcome of this iterative consensus process across all three test equations. All three \acp{PDE} achieve perfect structural accuracy, matching their respective governing equations exactly. The coefficient standard deviations consistently remain within acceptable ranges for all test cases, with values spanning from $2.47 \times 10^{-9}$ to $4.26 \times 10^{-8}$.

These results demonstrate that regression quality shows no direct correlation with numerical discretisation fineness or equation complexity. Despite the heat diffusion equation benefiting from the finest temporal resolution and simplest single-term structure, coefficient stability matches that of the more complex multi-term equations. Similarly, the \ac{KdV}-Burgers equation achieves reliable coefficient identification despite operating with coarser discretisation and managing three distinct physical terms simultaneously.

This demonstrates that the ensemble sparse regression approach remains robust to variations in numerical setup and equation complexity, provided the underlying problem remains well-posed. The consistent recovery of physically meaningful terms across all test cases indicates that the effectiveness of the methodology depends more fundamentally on the mathematical conditioning of the sparse regression problem than on the specific discretisation parameters or the number of terms involved.
\begin{table}[htbp]
\centering
\caption{Ensemble PDE identification showing complete equations and coefficient stability (standard deviation across bootstrap samples).}
\label{tab:pde_recovery}
\footnotesize
\begin{tabular}{@{}p{2.5cm}p{7cm}p{4cm}@{}}
\toprule
\textbf{PDE} & \textbf{Discovered Equation} & \textbf{Coeff Std Dev} \\
\midrule
Heat & $u_t = \nu u_{xx|cd2}$ & $\nu u_{xx|cd2}$: $2.6467 \times 10^{-8}$ \\
\midrule
Burgers & $u_t = -u u_{x|uw2} + \nu u_{xx|cd2}$ & $u u_{x|uw2}$: $2.4700 \times 10^{-9}$ \newline $\nu u_{xx|cd2}$: $1.9259 \times 10^{-8}$ \\
\midrule
\ac{KdV}-Burgers & $u_t = -C_1 u u_{x|uw2} - C_2 u_{xxx|cd2} + \nu u_{xx|cd2}$ & $C_1 u u_{x|uw2}$: $1.6212 \times 10^{-8}$ \newline $C_2 u_{xxx|cd2}$: $4.2591 \times 10^{-8}$ \newline $\nu u_{xx|cd2}$: $2.3253 \times 10^{-8}$ \\
\bottomrule
\end{tabular}
\end{table}

\subsection{Method Ablation Study}
% METHOD: Systematic comparison across method combinations  
% BASELINE: Standard SINDy with full library
% VARIANTS: +DSF only, +Gram only, +DSF+Gram, +Ensemble, +All
% TASK: Multiple PDEs across 100 parameter realisations each
% SHOW: Library size reduction, memory scaling, discovery accuracy
% METRICS: Library terms (before/after), memory usage (GB), coefficient recovery
% DEMONSTRATE: Each component (DSF, Gram, Ensemble) provides measurable benefit

The analysis systematically evaluates each pipeline component through controlled ablation experiments across 100 parameter realisations per PDE, isolating the specific benefits of \ac{DSF}, Gram accumulation, and ensemble consensus on both computational efficiency and identification accuracy.

All experiments employ identical discretisation and library construction procedures from Table~\ref{tab:case_studies_setup}, with only specified method components enabled.

Table~\ref{tab:ablation_results} demonstrates progressive improvements from each component. Gram accumulation (A2) reduces memory by 95\% while maintaining exact recovery. \ac{DSF} (A3) cuts library size from 22 to 17 terms. Their combination (A4) achieves optimal performance with exact identification in 14.5 seconds using only 0.04GB memory.

The \ac{KdV}-Burgers results reveal the critical role of dimensional filtering. Experiment B1 demonstrates the computational burden of processing large libraries without dimensional filtering, requiring 232.5 seconds and identifying 194 spurious terms from the 346-term library. \ac{DSF} transforms this scenario (B2), reducing the library to 124 terms and achieving exact 3-term identification in 101.2 seconds with 1.41GB memory consumption.

Particularly revealing, the comparison between experiments B3 and B4 demonstrates the critical interaction between ensemble methods and dimensional filtering. Experiment B3 applies ensemble consensus to the full 346-term library, yet achieves only partial identification with 5 active terms despite the computational expense of 2134.7 seconds and 34.7GB memory consumption. In contrast, experiment B4 combines ensemble methods with \ac{DSF}, achieving exact 3-term identification in 123.4 seconds, confirming that dimensional filtering proves essential for robust ensemble performance.

\begin{table}[htbp]
\centering
\caption{Ablation study results across method combinations without iterative pruning}
\label{tab:ablation_results}
\begin{threeparttable}
\resizebox{\textwidth}{!}{%
\begin{tabular}{llcccccccc}
\toprule
\textbf{Exp.} & \textbf{PDE} & \textbf{DSF} & \textbf{Gram} & \textbf{Ens.} & \textbf{Runtime (s)} & \textbf{RAM (GB)}\tnote{a} & \textbf{Active Terms} & \textbf{Outcome} \\
\midrule
A1 & Burgers & -- & -- & -- & 17.7 & 3.74 & 2/22 & Exact, baseline \\
A2 & Burgers & -- & \checkmark & -- & 16.1 & 0.19 & 2/22 & Exact, efficient \\
A3 & Burgers & \checkmark & -- & -- & 15.0 & 2.83 & 2/17 & Exact, leaner \\
A4 & Burgers & \checkmark & \checkmark & -- & \textbf{14.5} & \textbf{0.04} & 2/17 & Exact, optimal \\
\midrule
B1 & KdV-Burg. & -- & \checkmark & -- & 232.5 & 4.21 & 194/346 & Poor, high cost \\
B2 & KdV-Burg. & \checkmark & \checkmark & -- & \textbf{101.2} & \textbf{1.41} & 3/124 & Exact, efficient \\
B3 & KdV-Burg. & -- & \checkmark & \checkmark & 2134.7 & 34.7 & 5/346 & Partial, noisy \\
B4 & KdV-Burg. & \checkmark & \checkmark & \checkmark & 123.4 & 15.7 & 3/124 & Exact, robust\tnote{b} \\
\bottomrule
\end{tabular}%
}
\begin{tablenotes}[flushleft]
\small
\item[a] Peak memory usage during regression computation.
\item[b] \enquote{robust} indicates ensemble provides coefficient uncertainty bounds (see Table~\ref{tab:pde_recovery}).
\end{tablenotes}
\end{threeparttable}
\end{table}\textbf{}

The results illuminate the synergistic relationships between pipeline components. Gram accumulation dramatically reduces memory requirements while maintaining accuracy, enabling analysis of larger datasets. \ac{DSF} provides substantial computational savings by reducing library dimensionality.

Their combination achieves optimal efficiency for single-shot discovery and reveals a crucial insight: ensemble methods require \ac{DSF} for robust performance. Experiment B3 confirms this dependency, showing ensemble consensus failing without dimensional filtering, whereas B4 attains exact recovery when both components operate together. The elevated memory consumption in ensemble experiments reflects parallel processing, which practitioners can control by adjusting core count or using sequential fits.

\subsection{Symbolic Subgrid-Scale Closure Discovery} % Update the TABLE!!! (tol=0.25)
\label{subsec:symbolic_sgs}
% METHOD: Complete pipeline (DSF + Gram + Ensemble) 
% TASK: Discover SGS stress closure from coarse-grained Burgers' equation
% SHOW: Novel symbolic SGS model, comparison with Smagorinsky/DNN closures
% METRICS: SGS stress prediction accuracy, model interpretability, generalisation
% DEMONSTRATE: SINDy discovers physically meaningful closure models (main contribution)

Recent ML-based SGS modelling efforts have demonstrated the potential for data-driven approaches, though these typically rely on predetermined closure frameworks. Schmelzer et al. ~\cite{Schmelzer2020} developed \ac{SpaRTA} to discover corrections to the existing $k-\omega$ SST model through ElasticNet regression on predefined tensor polynomials. Kurz et al.~\cite{Kurz2023} employed \ac{RL} to learn spatially and temporally adaptive Smagorinsky coefficients for implicitly filtered LES. Both approaches assume specific closure structures a priori, whether linear eddy viscosity frameworks or Smagorinsky-type models, and focus on parameter optimisation within these known forms.

The \ac{SINDy} framework offers a fundamentally different approach by autonomously discovering the functional form of \ac{SGS} closures directly from data. Opaque models like deep neural networks might achieve high predictive accuracy. However, their lack of symbolic form makes direct physical interpretation challenging. They also complicate integration into existing \ac{CFD} equation sets. The primary objective of this work goes beyond maximising performance metrics. We aim to discover closures that are both accurate and fully interpretable. The framework preserves explicit symbolic forms without prior theoretical assumptions. Filtered flow fields govern $\tau_{SGS}$ behaviour. This study investigates whether \ac{SINDy} can directly discover \ac{SGS} stress closures from filtered velocity data.

The \ac{SINDy} pipeline employs box-filtered velocity fields $\bar{u}$, true \ac{SGS} stresses $\tau_{SGS}$, filter widths $\Delta$, and viscosities $\nu$ as input data. The regression datasets comprise velocity profiles with varying viscosities and corresponding box-filtered fields across filter widths $\Delta \in [1, 11]\delta x$, where $\delta x$ denotes the grid spacing. All remaining simulation parameters follow the specifications in Table~\ref{tab:case_studies_setup}.

Recognising that SGS closure discovery presents fundamentally different challenges compared to exact PDE identification, the implementation employs more stringent conditions. The analysis generates 500 parameter realisations across 50 ensemble batches with an adaptive coefficient of variation threshold $\text{CV}_{\text{threshold}} = \text{CV}_{\text{init}} \times (\text{decay rate})^{\text{iteration}}$, employing $\text{CV}_{\text{init}} = 0.5$ and decay rate of 0.5. The relaxed DSF tolerance of 0.25 enables the discovery of complex nonlinear dependencies whilst maintaining dimensional consistency. The framework constructs a polynomial library containing terms up to degree 3, including both $\Delta$ and $\Delta^2$ terms to capture multi-scale dependencies whilst reducing computational cost.

Through iterative stability-based pruning, the framework consistently identifies $\Delta^2 \bar{u}_{x}|_{cd1}^2$ as the dominant term across both regression methods. Given the convergence of both ElasticNet and SR3 to this identical functional form, we simplify the notation to $\bar{u}_{x}$ for clarity, yielding:
\begin{align}
\tau_{SGS} = C^{\text{SINDy}} \Delta^2 \bar{u}_{x}^2,
\label{eq:sindy_sgs_general}
\end{align}
where both methods converge to nearly identical coefficients ($C^{\text{ElasticNet}} = 0.1600$ vs $C^{\text{SR3}} = 0.1604$), with SR3 selected due to its lower coefficient variation:
\begin{align}
\tau_{SGS} = 0.1604 \Delta^2 \bar{u}_{x}^2.
\label{eq:sindy_sgs}
\end{align}

Table~\ref{tab:sgs_result} documents the detailed convergence analysis. Both methods achieve convergence to the single dominant term by iteration 7, with \ac{CV} dropping below 0.007, validating robust convergence across different regression approaches.

\begin{table}[htbp]
\centering
\caption{Iterative model convergence analysis showing stability-based term pruning process from initial 31-term library using filtered velocity fields.}
\label{tab:sgs_result}
\begin{threeparttable}
\resizebox{\textwidth}{!}{%
\begin{tabular}{@{}cllcccc@{}}
\toprule
\textbf{Method} & \textbf{Iteration} & \textbf{Active Terms} & \textbf{Selected Terms} & \textbf{Coeff.}\tnote{a} & \textbf{Std.}\tnote{a} & \textbf{CV}\tnote{a} \tnote{b} \\
\midrule
\multirow{7}{*}{\textbf{ElasticNet}} 
  & 1 & 13/31 
    & $\Delta^2 \bar{u}_{x|cd1}^2, \Delta^2 \bar{u}^2, \Delta^2 \bar{u}_{x|uw2}^2, \Delta \nu \bar{u}_{x|cd1}, \dots, \Delta^{-1}\bar{u}^2, \nu \bar{u}_{x|cd1}, \Delta^2 \bar{u} \bar{u}_{xx|cd2}$ 
    & 0.1937 & 7.31e-03 & 0.0377 \\
  & 2 & 10/13 
    & $\Delta^2 \bar{u}_{x|cd1}^2, \Delta^2 \bar{u}^2, \Delta^2 \bar{u}_{x|uw2}^2, \Delta \nu \bar{u}_{x|cd1}, \dots, \Delta^{-1}\bar{u}^2, \nu \bar{u}_{x|cd1}, \Delta^2 \bar{u} \bar{u}_{xx|cd2}$ 
    & 0.1862 & 2.84e-03 & 0.0152 \\
  & 3 & 6/10 
    & $\Delta^2 \bar{u}_{x|cd1}^2, \Delta^2 \bar{u}^2, \Delta^2 \bar{u}_{x|uw2}^2, \Delta \nu \bar{u}_{x|cd1}, \Delta^{-1}\bar{u}^2, \nu \bar{u}_{x|cd1}$ 
    & 0.1875 & 3.88e-03 & 0.0207 \\
  & 4 & 2/6 
    & $\Delta^2 \bar{u}_{x|cd1}^2, \Delta^2 \bar{u}^2$ 
    & 0.2029 & 5.21e-03 & 0.0257 \\
  & 5 & 1/2 
    & $\Delta^2 \bar{u}_{x|cd1}^2$ 
    & 0.1595 & 8.45e-04 & 0.0053 \\
  & 6 & 1/1 
    & $\Delta^2 \bar{u}_{x|cd1}^2$ 
    & 0.1601 & 9.30e-04 & 0.0058 \\
  & 7 & 1/1 
    & $\Delta^2 \bar{u}_{x|cd1}^2$ 
    & 0.1600 & 1.01e-03 & 0.0063 \\
\midrule
\multirow{7}{*}{\textbf{SR3}}
  & 1 & 9/31 
    & $\Delta^2 \bar{u}_{x|cd1}^2, \Delta^2 \bar{u} \bar{u}_{xx|cd2}, \Delta \nu \bar{u}_{x|cd1}, \Delta \nu \bar{u}_{x|uw2}, \dots, \Delta^{-1} \nu \bar{u}_{x|cd1}, \Delta^{-1} \nu \bar{u}_{x|uw2}$ 
    & 0.2436 & 1.09e-01 & 0.4486 \\
  & 2 & 9/9 
    & $\Delta^2 \bar{u}_{x|cd1}^2, \Delta^2 \bar{u} \bar{u}_{xx|cd2}, \Delta \nu \bar{u}_{x|cd1}, \Delta \nu \bar{u}_{x|uw2}, \dots, \Delta^{-1} \nu \bar{u}_{x|cd1}, \Delta^{-1} \nu \bar{u}_{x|uw2}$ 
    & 0.1563 & 2.78e-03 & 0.0178 \\
  & 3 & 8/9 
    & $\Delta^2 \bar{u}_{x|cd1}^2, \Delta^2 \bar{u} \bar{u}_{xx|cd2}, \Delta \nu \bar{u}_{x|cd1}, \Delta \nu \bar{u}_{x|uw2}, \dots, \Delta^{-1} \nu \bar{u}_{x|cd1}, \Delta^{-1} \nu \bar{u}_{x|uw2}$ 
    & 0.1579 & 3.47e-03 & 0.0220 \\
  & 4 & 4/8 
    & $\Delta^2 \bar{u}_{x|cd1}^2, \Delta^2 \bar{u} \bar{u}_{xx|cd2}, \Delta \nu \bar{u}_{x|cd1}, \Delta \nu \bar{u}_{x|uw2}$ 
    & 0.1706 & 1.20e-03 & 0.0070 \\
  & 5 & 1/4 
    & $\Delta^2 \bar{u}_{x|cd1}^2$ 
    & 0.1702 & 9.38e-04 & 0.0055 \\
  & 6 & 1/1 
    & $\Delta^2 \bar{u}_{x|cd1}^2$ 
    & 0.1598 & 7.66e-04 & 0.0048 \\
  & 7 & 1/1 
    & $\Delta^2 \bar{u}_{x|cd1}^2$ 
    & 0.1604 & 6.45e-04 & 0.0040 \\
\bottomrule
\end{tabular}%
}
\begin{tablenotes}[flushleft]
\small
\item[a] Coefficient value, standard deviation, and coefficient of variation for the dominant term $\Delta^2 \bar{u}_{x|cd1}^2$.
\item[b] Adaptive CV threshold with decay rate 0.5 progressively tightens stability requirements across iterations.
\end{tablenotes}
\end{threeparttable}
\end{table}

The highly consistent final coefficients across both regression methods confirm that this discovery represents the optimal single-term closure. In one dimension, the classical Smagorinsky stress exhibits odd parity in $\bar{u}_x$ and a negative sign, whereas the regression learns a magnitude-based dependence $\bar{u}_x^2$. To enforce the physically correct parity and sign while retaining the learned scaling, the analysis applies the signed variant:
\begin{align}
\tau_{SGS} = -C^{\text{SINDy}}\,\Delta^{2}\,\bar{u}_x\,|\bar{u}_x|
\label{eq:sindy_sgs_signed}
\end{align}
which corresponds to an effective Smagorinsky constant $C_s^{\text{SINDy}} = \sqrt{0.1604} \approx 0.4005$. All subsequent error analyses employ this signed form.

%%%%%%%%%%%%%%%%
The discovered constant $C_s^{\text{SINDy}} \approx 0.4005$ significantly exceeds $C_s \approx 0.16$~\cite{Li2016} for 3D isotropic turbulence due to fundamental physical differences. The classical Smagorinsky model accounts for energy dissipation from turbulent energy cascades~\cite{Pope2000}, while the 1D Burgers' equation dissipates energy primarily through sharp gradient formation and shocks.

SINDy discovers a larger coefficient to compensate for these structural differences—a phenomenon called \textit{model inadequacy}~\cite{Kennedy2001}. Grid-refinement analysis in Appendix \ref{sec:appendix_A} confirms this elevated coefficient persists across resolutions, demonstrating that SINDy correctly adapts the model constant to 1D shock-dominated physics rather than 3D turbulent cascades.
%%%%%%%%%%%%%%%%

\begin{figure}[!htbp]
    \centering
    \begin{subfigure}[t]{\textwidth}
        \centering
        \includegraphics[width=\textwidth]{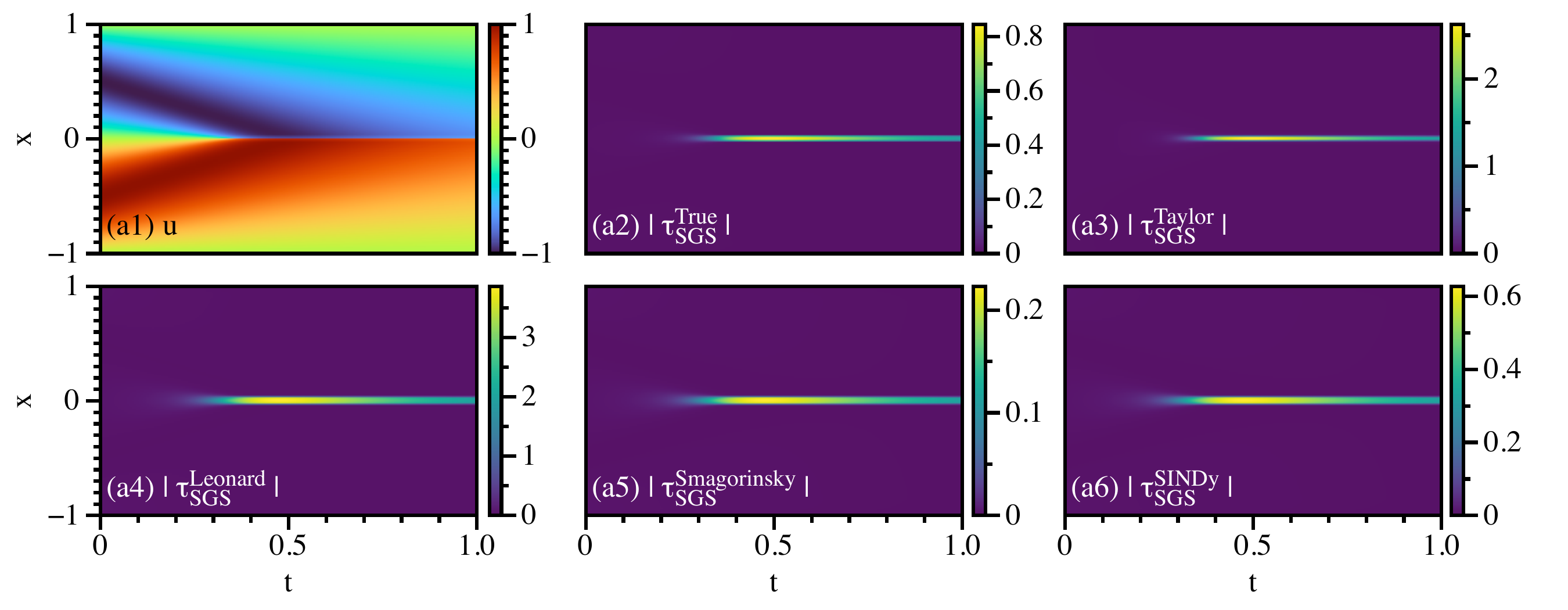}
        \caption{Sine wave initial condition $u_0 = -\sin(\pi x)$}
        \label{fig:sgs_sine}
    \end{subfigure}
    \\[0.5cm]
    \begin{subfigure}[t]{\textwidth}
        \centering
        \includegraphics[width=\textwidth]{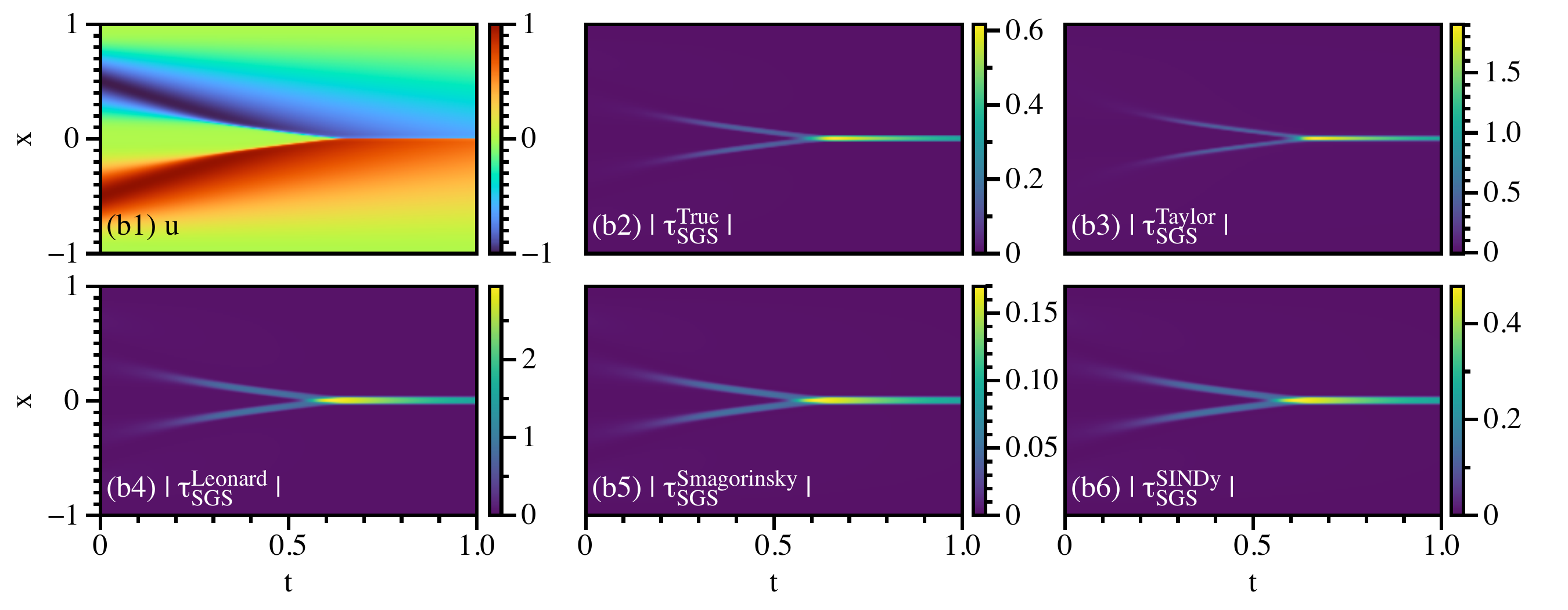}
        \caption{Cubic sine wave initial condition $u_0 = -\sin^3(\pi x)$}
        \label{fig:sgs_sine_cube}
    \end{subfigure}
    \caption{%
    SGS stress comparison for two analytical initial conditions. 
    Subfigure (a) corresponds to the sine wave initial condition $u_0 = -\sin(\pi x)$ 
    and subfigure (b) to the cubic sine wave initial condition $u_0 = -\sin^3(\pi x)$. 
    Each subfigure contains six panels showing: 
    (1) numerical Burgers' solution, 
    (2) true SGS stress, 
    (3) Taylor SGS model, 
    (4) Leonard SGS model, 
    (5) Smagorinsky SGS model, and 
    (6) SINDy SGS closure. 
    Panels (2)–(6) display absolute values for visual clarity, whereas quantitative analyses use signed stresses 
    as defined in Equation~\ref{eq:sindy_sgs_signed}. 
    The viscosity is fixed at $\nu = 0.005/\pi$ and the filter width at $\Delta_{\text{box}} = 5 \delta x$.
    }
    \label{fig:sgs_combined}
\end{figure}

%Update this part sounds bad
Figure~\ref{fig:sgs_combined} demonstrates SGS stress spatial distributions for sine wave and cubic sine wave initial conditions with viscosity $\nu = 0.005/\pi$ and filter width $\Delta_{\text{box}} = 5 \delta x$. Each subfigure compares the numerical Burgers' solution, true SGS stress, and four closure models (Taylor, Leonard, Smagorinsky, and SINDy). The SINDy-discovered model closely matches the true SGS stress structure across both configurations, particularly in high-gradient regions, while traditional models exhibit varying accuracy with notable deviations from the reference distribution.

\begin{figure}[htbp]
    \centering
    \begin{subfigure}[t]{0.49\textwidth}
    \centering
        \includegraphics[width=\textwidth]{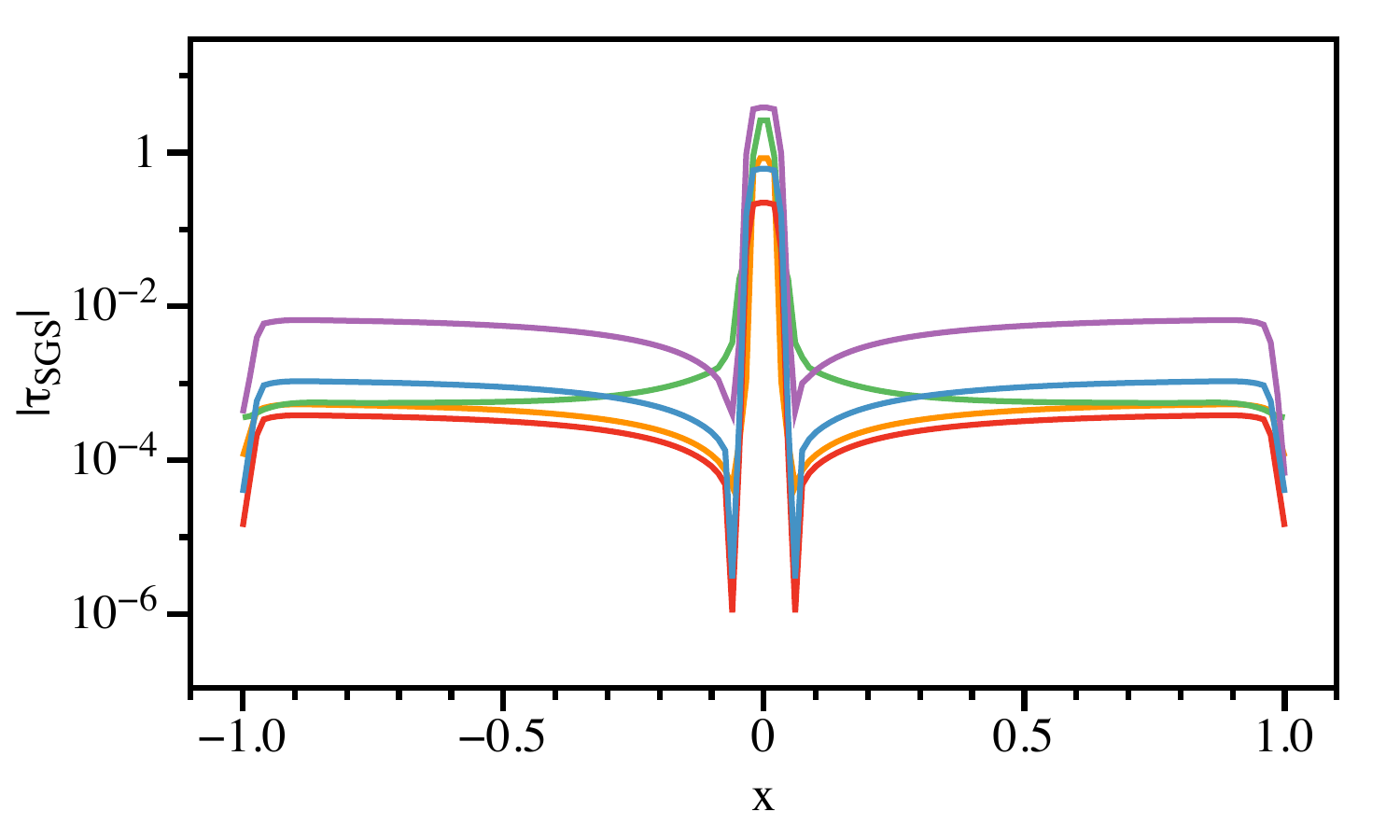}
        \caption{Sine wave initial condition}
        \label{fig:sgs_comparison_sine_t050}
        \end{subfigure}
        \hfill
        \begin{subfigure}[t]{0.49\textwidth}
    \centering
        \includegraphics[width=\textwidth]{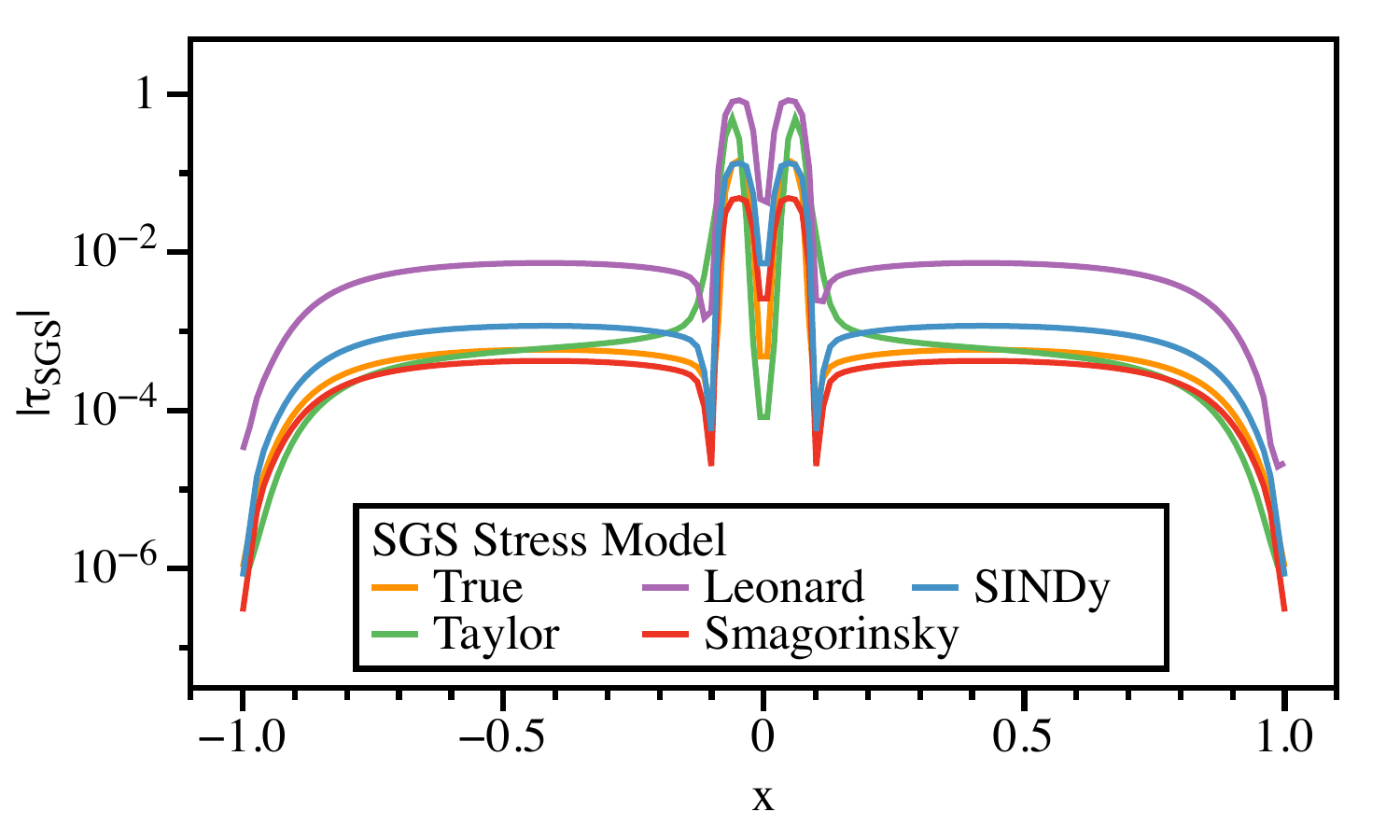}
        \caption{Cubic sine wave initial condition}
        \label{fig:sgs_comparison_sine_cube_t050}
        \end{subfigure}
        \caption{%
        SGS stress comparison at $t=0.50$ for 
        (a) $u_0=-\sin(\pi x)$ and (b) $u_0=-\sin^3(\pi x)$. 
        Stress magnitudes are plotted to emphasise structural differences, 
        while all evaluations use signed stresses (Eq.~\ref{eq:sindy_sgs_signed}). 
        The SINDy SGS closure remains closer to the true SGS stress in both cases, 
        whereas classical SGS models show larger discrepancies, particularly for the cubic initial condition.
        }
\label{fig:sgs_comparison_t050}
\end{figure}

Figure~\ref{fig:sgs_comparison_t050} provides point-wise comparisons at $t = 0.50$, confirming the agreement of SINDy with true SGS stress across the entire domain for both initial conditions. Traditional models show systematic deviations, particularly in moderate gradient regions, demonstrating that the discovered model generalises effectively across different flow configurations.

\begin{figure}[htbp]
    \centering
    \begin{subfigure}[t]{0.49\textwidth}
    \centering
        \includegraphics[width=\textwidth]{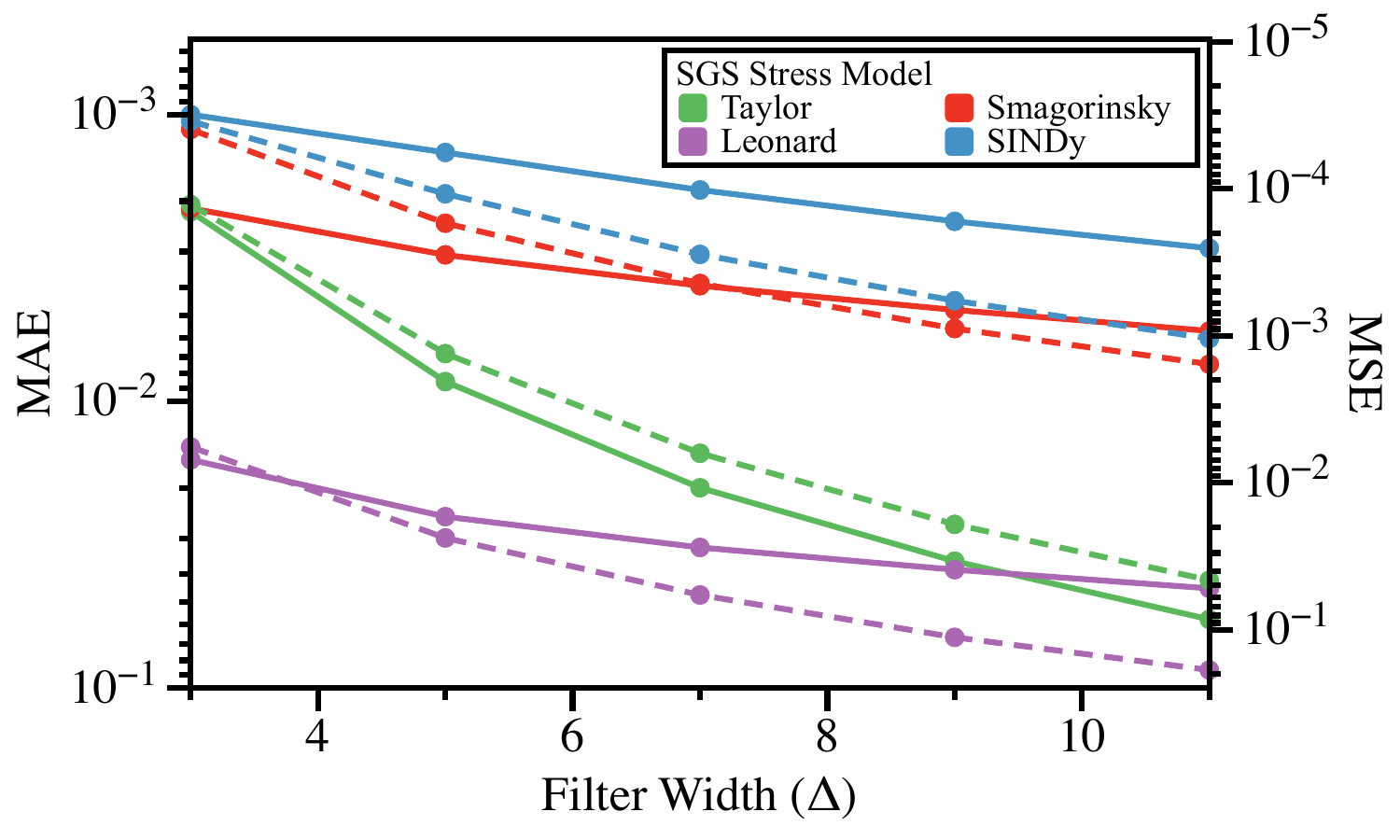}
        \caption{Comparison of all SGS stress models}
        \label{fig:performance_metric_large}
        \end{subfigure}
        \hfill
        \begin{subfigure}[t]{0.49\textwidth}
    \centering
        \includegraphics[width=\textwidth]{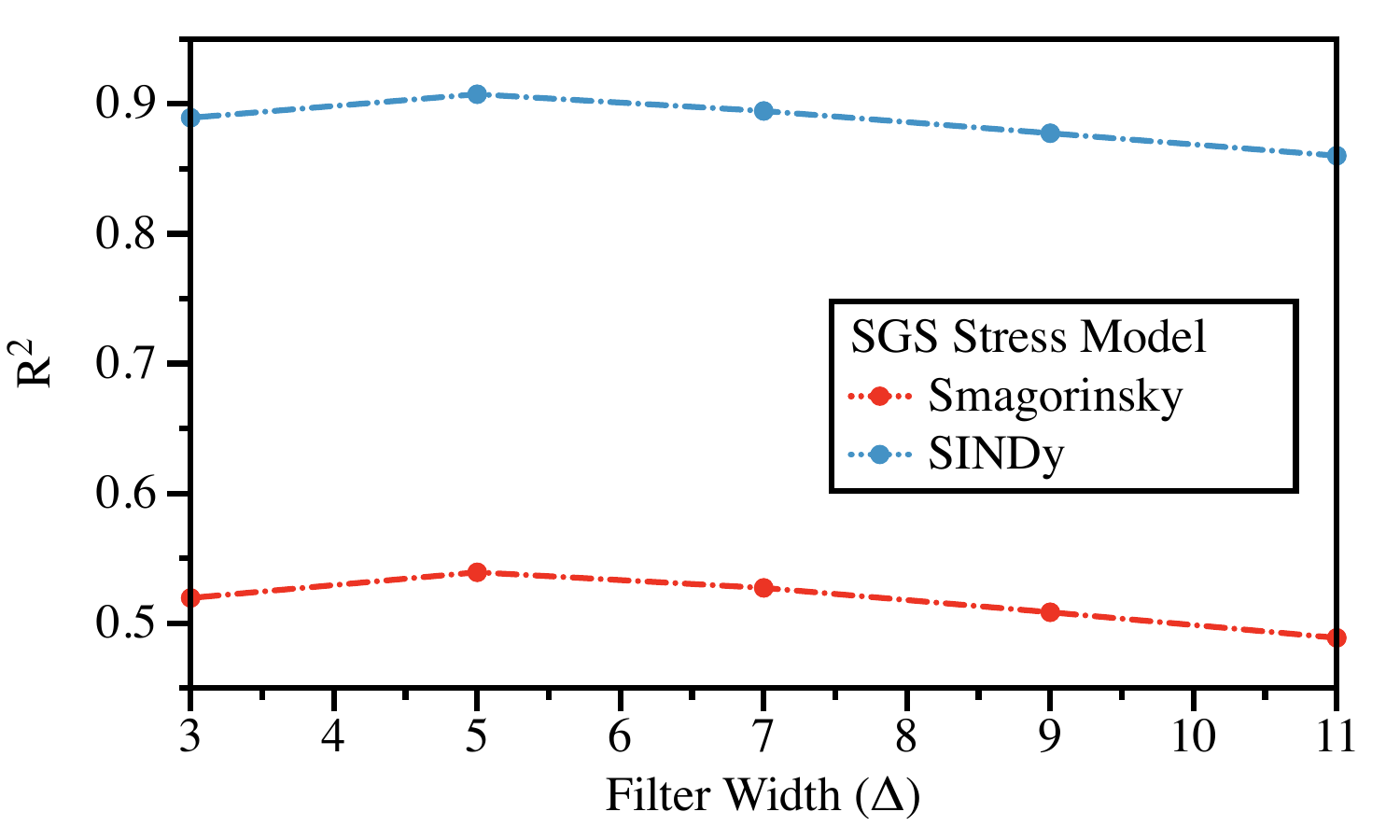}
        \caption{Comparison of Smagorinsky and SINDy}
        \label{fig:performance_metric_small}
        \end{subfigure}
        \caption{%
        Performance of SGS stress models across filter widths. 
        Subfigure (a) shows mean squared error (MSE, solid) and mean absolute error (MAE, dashed) 
        for all four models, while subfigure (b) reports $R^2$ scores for Smagorinsky and SINDy. 
        All metrics are evaluated on signed stresses. 
        Results are averaged over 50 randomised cases with Perlin noise initial conditions 
        and viscosities sampled from $\nu \in [0.001/\pi, 0.01/\pi]$. 
        The SINDy SGS closure maintains higher accuracy across filter widths, 
        whereas Taylor and Leonard models yield negative $R^2$ values, 
        highlighting their limited predictive capability.
        }
\label{fig:performance_metrics}
\end{figure}

Figure~\ref{fig:performance_metrics} presents quantitative assessment across multiple filter widths using 50 randomly generated Perlin noise initial conditions with viscosities sampled from $\nu \in [0.001/\pi, 0.01/\pi]$. The SINDy model consistently achieves the lowest error metrics with averaged $R^2 = 0.885$, while Taylor and Leonard models show negative $R^2$ values, highlighting their limited predictive capability.

The symbolic regression approach successfully discovered a physically meaningful subgrid-scale closure that autonomously reproduces the fundamental Smagorinsky structure while achieving superior predictive accuracy. The discovered model captures essential gradient-based physics without prior theoretical knowledge and maintains computational simplicity through a single-term structure, demonstrating the potential of SINDy to bridge the gap between manually tuned empirical models and black-box neural networks by providing physics-informed closure models with explicit functional forms compatible with existing CFD workflows.

\section{Conclusions and Future Work}

This work suggests an enhanced \ac{SINDy} pipeline designed to address key limitations in sparse regression for physics discovery. Traditional \ac{SINDy} implementations face computational constraints when handling large candidate libraries and typically require fixed physical parameters, restricting generalisability across different regimes.

The pipeline incorporates systematic solutions through \ac{DSF} and Gram matrix-based batch processing. The \ac{DSF} efficiently prunes physically inconsistent candidates while maintaining comprehensive exploration. Gram matrix decomposition reduces memory requirements by up to 95\% while preserving regression accuracy, enabling physical properties to be parameterised as symbolic variables rather than fixed constants.

Discovery of the \ac{SGS} stress for Burgers' equation illustrates the potential of the method. The framework identified $\tau_{SGS} \approx 0.1604 \Delta^2 \bar{u}_{x}^2$, corresponding to a Smagorinsky constant of about 0.4005, without prior assumptions about closure form. The model achieved strong predictive performance with $R^2 = 0.885$ across filter widths, outperforming classical Taylor and Leonard closures.

The approach complements existing ML-based turbulence modelling efforts. Whereas recent methods such as \ac{SpaRTA} enhance predetermined frameworks and \ac{RL} optimise parameters within known structures ~\cite{Schmelzer2020}, the symbolic regression presented here discovers functional relationships directly from data while retaining interpretability. This offers an alternative pathway for closure development that balances automation with physical insight.

The framework enables symbolic representation and direct comparison between discovered and established models, providing data-driven insight into closure constants. The method thereby fulfils the four primary objectives of parameter-aware libraries, memory-efficient regression, physics-constrained selection, and symbolic \ac{SGS} discovery.

This work contributes a valuable tool to the expanding set of data-driven methodologies in computational fluid dynamics by demonstrating the ability to discover turbulence closure relationships while retaining physical interpretability. Future work will focus on extending the framework to a range of higher-dimensional problems, from 2D and 3D turbulent flows to more complex physical scenarios such as reacting flows. Further research avenues include the exploration of adaptive or scale-dependent closures and the development of modules for seamless, \textit{plug-and-play} integration with conventional CFD solvers.

% System Info
%Python version: 3.11.13
%numpy version: 2.0.2
%pandas version: 2.2.2
%scipy version: 1.16.0
%scikit-learn version: 1.6.1
%perlin-noise version: 1.2.1
%v2-8 TPU
% Seed = 1234

\bibliographystyle{unsrt}
\bibliography{references}

\newpage
\begin{appendices}

\section{Grid refinement analysis of SGS coefficients}
\label{sec:appendix_A}

Section \ref{subsec:sgs_stress} demonstrated that when the library contained multiple candidate terms, repeated runs of the pipeline consistently identified $\Delta^2 \bar{u}_x^2$. This outcome aligns with the Smagorinsky closure, although it reflects a data-driven regression sensitive to the chosen configuration. The Appendix, therefore, shifts focus. It assumes the Smagorinsky form, restricts the library to $\Delta^2 \bar{u}_x^2$, and evaluates how the corresponding coefficient changes with grid refinement.

The analysis explores the effective constant as a function of grid spacing $\delta x$. Running the pipeline with a single candidate term at each resolution produces coefficients that enable direct comparison across discretisations.

\begin{figure}[!htbp]
\centering
\includegraphics[width=0.55\textwidth]{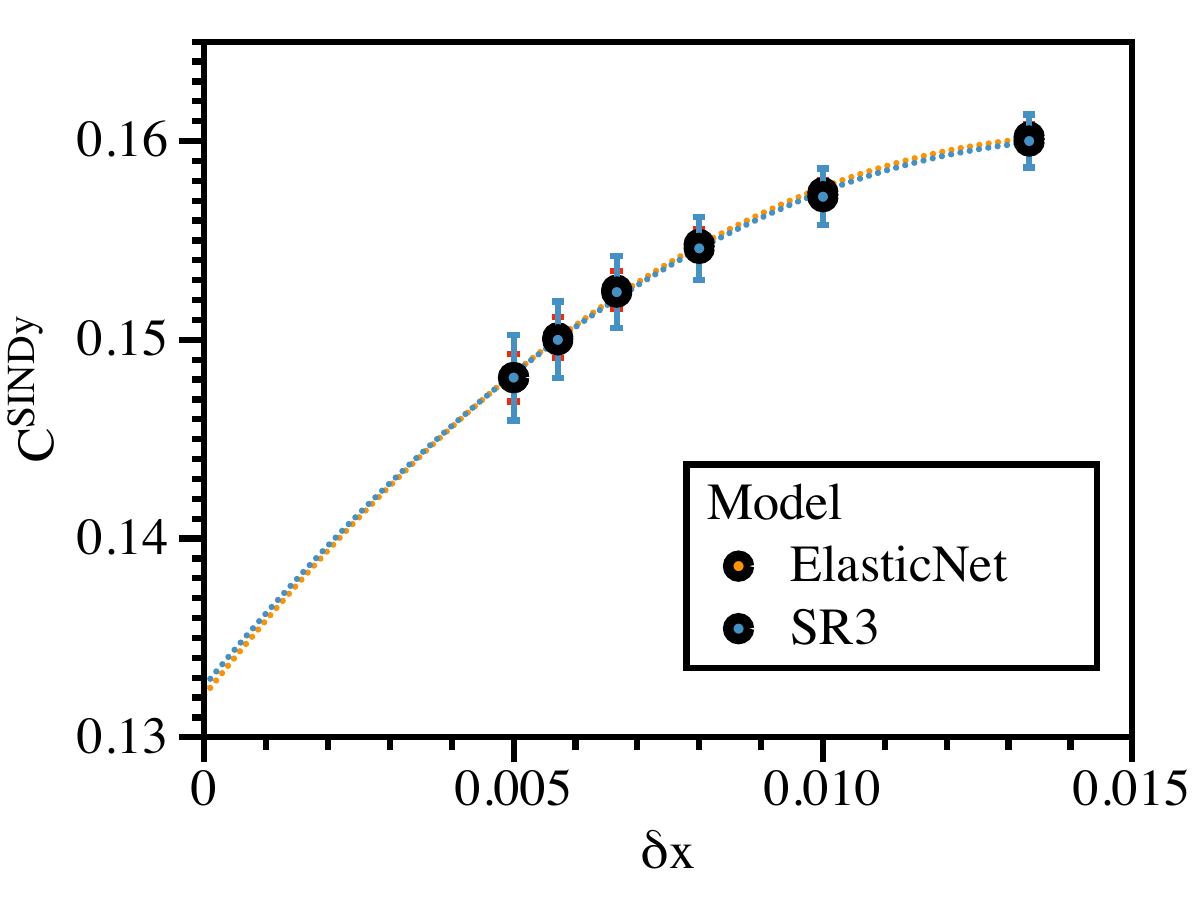}
\caption{Variation of the effective SINDy regression coefficient $C^{\text{SINDy}}$ with grid spacing $\delta x$, obtained using the same SINDy pipeline with a restricted library. Circles denote regression results, while the dashed lines indicate quadratic fits. ElasticNet and SR3 regressions produce nearly identical trends.}
\label{fig:cs_change}
\end{figure}

Figure~\ref{fig:cs_change} and Table~\ref{tab:cs_refinement} report the outcomes. The effective coefficient $C^{\text{SINDy}}$ decreases from 0.1602 to 0.1481 as the grid is refined, consistent with the truncation-error scaling of central-difference operators. ElasticNet and SR3 regressions converge on nearly identical values, reinforcing the robustness of the observed trend. Minor deviations at $\delta x=0.0133$ reflect well-documented sensitivities of data-driven regression~\cite{Brunton2016, Champion2019, Hastie2009}, with the single-run appendix result differing slightly from the ensemble-averaged values reported in the main text. These deviations remain confined to the third decimal place and preserve the physical interpretation.

\begin{table}[htbp]
\centering
\caption{Grid-refinement study of the effective Smagorinsky constant $C_s$. ElasticNet and SR3 regressions produce effectively identical results.}
\label{tab:cs_refinement}
\small
\begin{tabular}{@{}c|ccc|ccc@{}}
\toprule
\multirow{2}{*}{\textbf{$\delta x$}} & \multicolumn{3}{c|}{\textbf{ElasticNet}} & \multicolumn{3}{c}{\textbf{SR3}} \\
\cmidrule{2-7}
 & $C_s$ & $C_s^2$ & std. & $C_s$ & $C_s^2$ & std. \\
\midrule
0.0133 & 0.4003 & 0.1602 & 6.53e-4 & 0.4000 & 0.1600 & 1.33e-3 \\
0.0100 & 0.3967 & 0.1574 & 6.58e-4 & 0.3965 & 0.1572 & 1.42e-3 \\
0.0080 & 0.3935 & 0.1548 & 7.80e-4 & 0.3932 & 0.1546 & 1.58e-3 \\
0.0067 & 0.3905 & 0.1525 & 9.49e-4 & 0.3904 & 0.1524 & 1.82e-3 \\
0.0057 & 0.3874 & 0.1501 & 1.03e-3 & 0.3873 & 0.1500 & 1.93e-3 \\
0.0050 & 0.3848 & 0.1481 & 1.20e-3 & 0.3848 & 0.1481 & 2.16e-3 \\
\bottomrule
\end{tabular}
\end{table}

\clearpage
\section{Dimensional Similarity Filter Algorithm}

\begin{algorithm}[H]
\caption{Dimensional Similarity Filter (DSF) for 1D Problems}\label{alg:dsf}
\begin{algorithmic}[1]
\State \textbf{Input:} Raw library $\Theta_{\text{deg}}$, target dimensions $\mathbf{d}^{\star} = [d_L^{\star}, d_T^{\star}]$, filter type $T_{filter}$, tolerance $\delta_{\text{tol}}$
\State \textbf{Output:} Filtered library $\Theta_{\text{DSF}}$
\State
\State Initialise $\Theta_{\text{DSF}} \leftarrow \emptyset$ and base dimension vectors for all primitive quantities
\If{$T_{filter}$ is `soft'}
    \State Compute $\max_L \leftarrow \max_{\theta \in \Theta_{\text{deg}}} |d_\theta^L|$ and $\max_T \leftarrow \max_{\theta \in \Theta_{\text{deg}}} |d_\theta^T|$
\EndIf
\For{each term $\theta \in \Theta_{\text{deg}}$}
    \State Compute dimension vector $\mathbf{d}_\theta = [d_\theta^L, d_\theta^T]$ by summing primitive dimensions
    \If{$T_{filter}$ is `hard'}
        \State $\textbf{pass} \leftarrow (\mathbf{d}_\theta = \mathbf{d}^{\star})$
    \ElsIf{$T_{filter}$ is `soft'}
        \State $d_L \leftarrow |d^{\star}_L - d^L_\theta| / \max_L$, $d_T \leftarrow |d^{\star}_T - d^T_\theta| / \max_T$
        \State $\delta \leftarrow (d_L + d_T) / 2$ \Comment{Taxicab metric}
        \State $\textbf{pass} \leftarrow (\delta \le \delta_{\text{tol}})$
    \Else
        \State $\textbf{pass} \leftarrow \textbf{true}$
    \EndIf
    \If{\textbf{pass}} Add $\theta$ to $\Theta_{\text{DSF}}$ \EndIf
\EndFor
\State \textbf{return} $\Theta_{\text{DSF}}$
\end{algorithmic}
\end{algorithm}

\end{appendices}

\end{document}